\pdfoutput=1

\documentclass[11pt]{article}

\usepackage{EMNLP2022}

\usepackage{times}
\usepackage{latexsym}

\usepackage[T1]{fontenc}

\usepackage[utf8]{inputenc}
\usepackage{microtype}

\usepackage{inconsolata}

\usepackage{physics}
\usepackage{mathtools,nccmath}
\usepackage{graphicx}
\usepackage{booktabs}
\usepackage{amsfonts,amssymb}
\usepackage{bm}
\usepackage{threeparttable}
\usepackage{multirow}

\usepackage{amsmath}
\usepackage{arydshln}
\usepackage{subfigure}
\usepackage{enumitem}
\usepackage{times}
\usepackage{soul}
\usepackage{url}
\usepackage{amsthm}
\usepackage{booktabs}
\usepackage{algorithm}
\usepackage{algorithmic}
\usepackage{xcolor}

\usepackage{caption}

\usepackage{color}

\title{ Learning Semantic Textual Similarity via Topic-informed \\ Discrete Latent Variables}

\author{Erxin Yu\textsuperscript{\rm 1,2}, Lan Du\textsuperscript{\rm 4}, Yuan Jin\textsuperscript{\rm 4}, Zhepei Wei\textsuperscript{\rm 1,2},  Yi Chang\textsuperscript{\rm 1,2,3\thanks{~~Corresponding Author}}\\ 
 \textsuperscript{\rm 1}School of Artificial Intelligence, Jilin University\\
 \textsuperscript{\rm 2}Key Laboratory of Symbolic Computation and Knowledge Engineering, Jilin University\\
 \textsuperscript{\rm 3}International Center of Future Science, Jilin University\\
 \textsuperscript{\rm 4}Faculty of Information Technology, Monash University, Australia\\
 erxin.yu@outlook.com,
 lan.du@monash.edu, \\
 yuan.jin@monash.edu,
 weizp19@mails.jlu.edu.cn,
 yichang@jlu.edu.cn\\
}

\begin{document}
\maketitle
\begin{abstract}
     Recently, discrete latent variable models have received a surge of interest in both Natural Language Processing (NLP) and Computer Vision (CV), attributed to their comparable performance to the continuous counterparts in representation learning, while being more interpretable in their predictions.~In this paper, we develop a topic-informed discrete latent variable model for semantic textual similarity, 
     which learns a shared latent space for sentence-pair representation via vector quantization.
    Compared with previous models limited to local semantic contexts, our model can explore richer semantic information via topic modeling.
    We further boost the performance of semantic similarity by injecting the quantized representation into a transformer-based language model with a well-designed semantic-driven attention mechanism.~We demonstrate, through extensive experiments across various English language datasets, that our model is able to surpass several strong neural baselines in semantic textual similarity tasks.
\end{abstract}

\section{Introduction}
    Semantic Textual Similarity (STS), which concerns the problem of measuring and scoring the relationships or relevance of pairs of text on real-valued scales, is
    a fundamental task in NLP. 
    It has further driven many other important NLP tasks such as machine translation~\cite{sts-ml}, text summarization~\cite{sts-ts}, question answering~\cite{sts-qs}, etc.

    Recently, deep neural language models have achieved state-of-the-art performance for STS. The success of these models is attributed to their adoption of self-supervised learning on text representations, which overcomes the absence of large labeled STS data in many domains. The text representations learned by these models, 
    usually consist of continuous latent variables, and can be fed pairwise into some functions (e.g. a multi-layer perceptron) to compute their semantic similarity. 

    Compared to the continuous ones, discrete latent variables have drawn much less attention in language modeling, despite that natural languages are discrete in nature, and there is growing evidence, from several NLP tasks \cite{vq-clssification,bao-etal-2020-plato}, that they are equally good, if not more suitable, as the continuous counterparts. As a widely-used training technique for learning discrete latent variables, vector-quantized variational autoencoder (VQ-VAE)~\cite{VQ-VAE} computes the values for the variables through the nearest neighbor look-up across the quantized vectors from the shared latent embedding space. Despite its success in speech recognition and computer vision, VQ-VAE has yet to be investigated in its use and viability in general NLP tasks. In this paper, we intend to explore the use of VQ-VAE and its generated discrete latent variables 
    for semantic textual similarity.
    
    Two issues need to be resolved in order for the VQ-VAE to work effectively on the STS task. First, the codebook embeddings need to be carefully initialized to prevent the so-called codebook collapse, in which only a few of the embeddings are selected and learned by the model, causing a reduction in the representational capacity of the codebook. Past research has indicated that sufficiently informed and sophisticated manipulation of the embedding initialization can mitigate this problem. A second issue is that most STS scenarios are designed to teach a language model to measure the textual similarity from a local perspective (e.g. over contexts within each sentence). However, global semantics underlying a broader context has already been shown to benefit a number of NLP tasks, including language generation \cite{language-generation}, textual similarity calculation ~\cite{tbert}, keyphrase generation ~\cite{keyphrase}, etc. For STS, it can provide perspectives on the global correlations and dependencies among sentences, which can be leveraged to better distinguish the semantic nuances of sentence-level contexts.
    
    In this paper, we propose to leverage topic modeling to tackle both issues. 
    The topic information 
    can provide informative guidance on the initialization and the learning of VQ-VAE's codebook.
    Furthermore, since the discovered topics can capture global semantics 
    (e.g., semantics as distributions over the vocabulary shared by the whole corpus), 
    they can be used to calibrate the bias of local semantics on the measure of textual similarity. 
    
    Our proposed topic-enhanced VQ-VAE language model features two major components to enrich the codebook representational capacity with rich contextual information.
    One is the topic-aware sentence encoder, 
    which computes the sentence embedding using its (local) topic distribution and the corresponding global topic-word distributions. 
    The other is a topic-guided VQ-VAE codebook, 
    where we align its latent codes/embeddings with the topic embeddings jointly learned by an NTM throughout the training phase. 
   We further incorporate the quantized sentence representations learned by our topic-enhanced VQ-VAE into transformer-based language models to guide the learning of contextual embeddings.
   Built on top of the multi-head attention, 
    a semantics-driven attention mechanism is proposed to compute the attention scores based on the quantized sentence representations. The proposed attention mechanism is a flexible plug-and-play module that requires a small number of extra parameters and a minimum change to the existing multi-head attention implementation. 
    Our contributions are summarised as:
        \begin{itemize}[topsep=0pt, partopsep=0pt, noitemsep, leftmargin=*]
            \item  To the best of our knowledge, this is the first language model that explores the use of discrete latent variables learned by VQ-VAE for STS.
            \item We introduce a topic-enhanced VQ-VAE model that combines the merits of topic modeling and vector quantization, where the quantization is informed by the topic information jointly learned by an NTM to enhance its robustness against the collapsing problem and its capacity of capturing both the global and local semantics. 
            \item We present a simple yet effective semantics-driven attention mechanism to inject the quantized representations into transformer models, which can serve as a plug-and-play module for almost all the existing transformer-based models.
            \item Comprehensive experiments on six real-world datasets for semantic textual matching demonstrate the effectiveness of our model and the significance of its major components.
        \end{itemize}

\section{Related Work}
\subsection{Neural Language Models for STS}
    Neural language models have played a key role in fulfilling the STS task. Among them, the earlier CNN-based model~\cite{cnn1,abcnn} and the RNN-based model~\cite{rnn1} used to achieve the state-of-the-art performance
    before the emergence of
    the transformer architecture~\cite{transformer} and the large-scale pre-training of its corresponding models. These pre-trained models, with their ability to be fine-tuned on various domains and tasks, have dominated the entire NLP area, significantly improving the task performance in the specific domains. As the most prominent model, BERT~\cite{BERT}, along with its variants including RoBERTa~\cite{RoBERTa}, BERT-sim~\cite{bert-sim}, ALBERT~\cite{albert} and SemBERT~\cite{semBert}, have achieved superior results in the STS task. Despite the achieved success, current language models are mostly ``black-box'' models with low interpretability, which require additional explanatory components for their predictions in needed fields such as Biomedicine and Finance. 
    
    
\subsection{Neural Topic Models}
    The success of Variational AutoEncoder (VAE) ~\cite{vae} has led to the development of a series of neural topic models (NTM)~\cite{wordEmbeddingsTopic,OptimalTransport}, (Refer to \citeauthor{ethan2021}~(\citeyear{ethan2021}) for a comprehensive survey on NTMs), 
    where the posterior distribution of document topics is approximated by a neural network during the variational inference process, known as the neural variational inference (NVI) \cite{NVDM}. The NVI also enables NTMs to be easily combined with various language models for capturing the global semantics information. These hybrid models are known as the topic-aware language models. Their combination strategies can be largely characterized by either concatenation of (document) topic distributions with the local word embeddings, or (multi-head) attention mechanisms towards the inferred topics. These models have been shown to achieve higher performance than the plain language models on various tasks, such as text summarization~\cite{friendly}, keyphrase generation ~\cite{keyphrase} and document classification~\cite{TMN}.
    
    \subsection{VQ-VAE}    
    Rather than learning continuous representations,
    VQ-VAE ~\cite{VQ-VAE} learns quantized representations via vector quantization.
    It was first applied to  speech representation~\cite{VQ-speech} and video generation~\cite{video-gen}.
    Recently, it has attracted lots of attention in a variety of NLP tasks. In machine translation, \citeauthor{vq-NMT}~(\citeyear{vq-NMT}) applied VQ-VAE to decode the target language sequence from the discrete space, making the model inference faster and the model performance only slightly compromised. In text classification, \citeauthor{vq-clssification}~(\citeyear{vq-clssification}) explored general text representations induced by VQ-VAE for the classification especially in low-resource domains. \citeauthor{vq-event}~(\citeyear{vq-event}) designed a semi-supervised VQ-VAE for new event type induction, while \citeauthor{vq-evidence}~(\citeyear{vq-evidence}) equipped the encode-decoder architecture with the VQ-VAE to automatically select relevant evidence. 
    In this paper, we move one step further by guiding the vector quantization with topics learned by NTMs.
    
\section{Modelling Framework}
\begin{figure*}[t]
        \centering
        \includegraphics[width=0.9\textwidth]{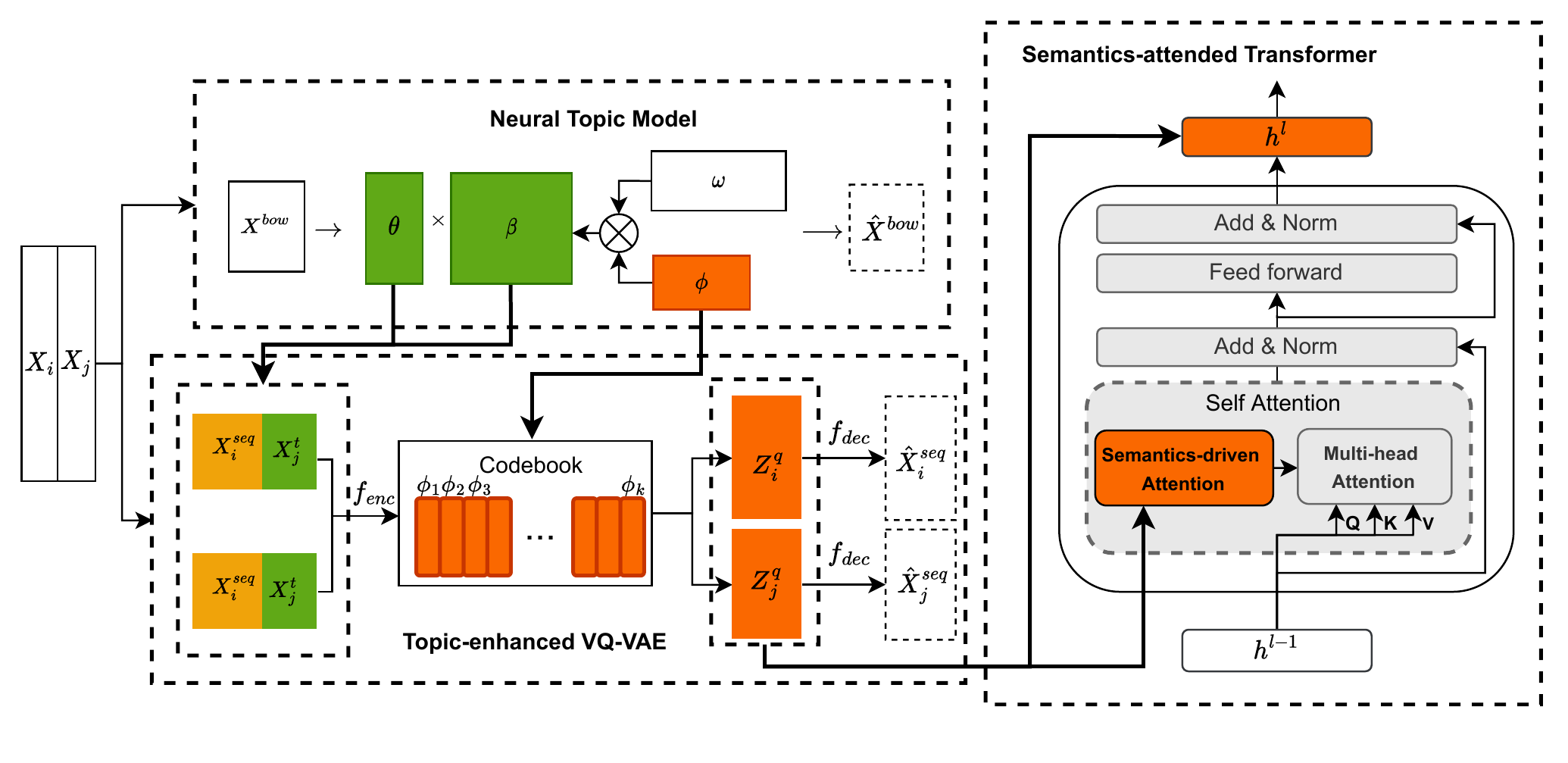}
        \vspace{-5mm}
        \caption{Our semantic textual matching framework combines the benefit of NTMs, VQ-VAE and Transformer.}
        \label{fig:Framework}
        \vspace{-1mm}
    \end{figure*}

    Given a pair of sentence $X = \left\{ \vb*{X}_i, \vb*{X}_j\right\}$ as the input,  
    where each sentence is represented as the bag-of-words (bow) vectors and contextualized embedding,
    our modelling framework, as shown in Figure \ref{fig:Framework},
    predicts the semantic similarity $Y$ between $\vb*{X}_i$ and $\vb*{X}_j$.
    
    
   
    \subsection{Neural Topic Component}   
    \label{NTM}
        We adopt a VAE-based neural topic model  \cite{NTM} to learn latent topics. 
        Different from LDA \cite{LDA}, 
        NTM parameterises the latent topics $\vb*{\theta}$ using a neural network conditioned on a draw from a Gaussian Softmax Construction. 
        Here, $\vb*{\theta} \in \mathbb{R}^K$ represents the topic proportions of a sentence $X$,  
        where $K$ denotes the number of topics. 
        Let $\vb*{\beta}_{k} \in \mathbb{R}^{V}$ be a distribution over a vocabulary $V$ associated with a topic $k$. 
        Following \cite{pmlr-v108-wang20c}, we use word embeddings $\vb*\omega \in \mathbb{R}^{V \times E}$, topic embeddings $\vb*\phi \in \mathbb{R}^{K \times E}$ to compute $\vb*{\beta}_{k}$ as
           $ \vb*{\beta}_{k} = Softmax(\vb*\omega \cdot \vb*{\phi}_{k}^{T})$.
        
         With NVI, we can use an encoder network to 
         approximate the true posterior.
         Specifically, the encoder generates the variational parameters $\vb*\mu$ and $\vb*\sigma^2$ through neural networks and  the latent variable $\vb*\theta$ is sampled with the Gaussian reparameterization trick. 
        The decoder reconstructs the BoW representation of a document by maximizing the log-likelihood of the input.
        The loss function (i.e., ELBO) contains a reconstruction error term and a KL divergence term as 
\setlength{\abovedisplayskip}{3pt}
\setlength{\belowdisplayskip}{3pt}
        \begin{equation*}\scriptstyle
            L_{NTM} = D_{KL}[q(\vb*\theta|\vb*{x})||p(\vb*\theta|\vb*{x})]
                     - E_{q(\vb*\theta|x)}[\sum_{m=1}^{M}\log p(w_m|\vb*\theta)],
        \end{equation*}     
        where $q(\vb*\theta|\vb*{x})$ denotes the variational posterior distribution of $\vb*\theta$ given the sentence $\vb*{x} = ( w_{1}, ..., w_{m})$, approximating the true posterior $p(\vb*\theta|x)$.
        
    \subsection{Topic-enhanced VQ-VAE} 
    \label{VQVAE}
        VQ-VAE \cite{VQ-VAE} takes the contextualized embeddings of words within a sentence: $\vb*{X}^{seq} = ( \vb*{x}_{1}^{seq}, ..., \vb*{x}_{m}^{seq})$ into its encoder to produce the corresponding latent representations $\vb*{Z}^e$. Then, the quantized representations $\vb*{Z}^q$ are calculated by the nearest neighbor look-up using the predefined and shared embedding space $\vb*{E}$, and further used as an input to the decoder to reconstruct the original sentence text. 
        
        However, since the original embedding space of VQ-VAE is randomly initialized, these learned embeddings can be arbitrary without clear semantic meaning.
        To guide the  codebook learning, we incorporate the topic information into VQ-VAE
        by designing the following 
        two components:
        
        \paragraph{Topic Sensitive Encoder.} 
        We use a standard single-layer Transformer~\cite{transformer} as our encoder.
        Specifically, for a sentence consisting of a sequence of $m$ words $\boldsymbol{X} = ( w_{1}, ..., w_{m})$, its corresponding word representations $\vb*{X}^{seq}$ are generated by the transformer embedding layer, which concatenates the word embedding and position embedding. 
        To incorporate the topic information, 
        we leverage the sentence topic distribution $\vb*\theta$ and the word topic weight vector $\vb*{\beta}_{w_m} \in \mathbb{R}^{K}$ from the NTM model to compute the topical embedding of the words $\vb*{X}^{t}  = \left\{ \vb*{x}_{1}^{t}, ..., \vb*{x}_{m}^{t} \right\} \in \mathbb{R}^{m \times K}$, where 
         $
            \vb*{x}_{m}^{t} \in \mathbb{R}^{K} = \vb*{\theta} \otimes  {\vb*{{\beta}}_{w_m}}.
        $
        We concatenate the $\vb*{X}^{seq}$ and $\vb*{X}^t$ and feed them into the encoder as
        $
            \vb*{Z}^{e} \in \mathbb{R}^{m\times E} = f_{enc}(\vb*{X}^{seq} \oplus \vb*{X}^{t}) .
        $
        \paragraph{Topical Latent Embedding.}
        We get final word representations $\vb*{Z}^{e} = \{ \vb*{z}_{1}^{e}, ..., \vb*{z}_{m}^{e} \}$ from the topic sensitive encoder. These representations are mapped onto the nearest element of embedding $\vb*{E}$. 
        Different from previous VQ-VAE models that use a randomly initialized embedding $\vb*{E}$, we leverage the topic embedding $\vb*\phi \in \mathbb{R}^{K\times E}$ from NTM. The discretization process is defined as: For each $i \in \{1, \dots, m\}$, $
                \vb*{z}_{i}^q = \vb*{\phi}_k \in \mathbb{R}^{E},\,\mbox{where}\, k = \mathop{\arg\min}_{j \in \left\{1,\dots,k\right\}} \left\| \vb*{z}_i^e - \vb*{\phi}_j \right\|_2.
            $
        Thus, we have quantized representations $\vb*{Z}^{q} = \left\{ \vb*{z}_{1}^{q}, ..., \vb*{z}_{m}^{q} \right\} \in \mathbb{R}^{m\times E}$ for a sentence.
        We then feed $\vb*{Z}^{q}$ into the decoder which is also a single-layer transformer to reconstruct the sentence X.
        The overall training objective is thus defined as
        
        {%
        \small
        \begin{align}
            L_{VQ-VAE} =& \sum_{i=1}^{m}(-\log p(\vb*{x}_{i}|\vb*{z}_{i}^{q}) + \left\| sg[\vb*{z}_{i}^e] - \vb*{z}_{i}^q \right\|_2^2\nonumber\\
            & + \lambda\left\| \vb*{z}_{i}^e - sg[\vb*{z}_{i}^q] \right\|_2^2).
        \end{align}   
         }%
        The first term above is the reconstruction error of the decoder given $\vb*{Z}^q$. The last two terms are used to 
        minimize the distance between the latent embedding $\vb*{E}$ and the encoder output $Z^e$,
        where $\vb*{\lambda}$
        is a commitment loss and $sg(\cdot)$ means the stop-gradient operation.
        It is noteworthy that our topic-enhanced VQ-VAE model jointly trained with NTM can explicitly learn
        the topic assignment of each individual words, which is not feasible for NTM alone.
        
    \subsection{Semantics-driven Multi-head Attention} 
    The multi-head attention in transformer-based models explores the relationships among tokens by calculating the token similarities, defined as 
    
    \begin{small}
    \begin{align}
        \vb*{Q} = &\vb*{K} = \vb*{V} = \vb*{h}^{l-1}\nonumber\\
        \text{MultiHead}(\vb*{Q},\vb*{K},\vb*{V})& = \text{Concat}(\vb*{head}_1,\cdots,\vb*{head}_n)\vb*{W}^{O}\nonumber\\
        \vb*{head}_i = \text{Attention}&(\vb{Q_i}{W}_i^{Q},\vb*{K_i}{W}_i^{K},\vb*{V_i}{W}_i^{V})
    \end{align}  
    \end{small}
    where $\vb*{h}^{l-1}\in \mathbb{R}^{L \times (n\cdot E)}$ is the output of last layer, $L$ is the length of sentence pair, n is the number of attention head, $E$ is the hidden size of each attention head. $\vb*{W}_i^{Q}$, $\vb*{W}_i^{K}$, $\vb*{W}_i^{V}$ and $\vb*{W}_i^{O} \in \mathbb{R}^{E \times E}$ are projection matrices. 
    Scaled Dot-Product method is adopted to calculate the attention function:
    
     {\small
    \begin{equation}
        {\text{Attention}(\vb*{Q},\vb*{K},\vb*{V})} = \text{Softmax}(\dfrac{\vb*{QK}^{T}}{\sqrt{d_k}})\vb*{V}
    \end{equation}
    \vspace{-4mm}
    }
    
    To enable the transformer-based model to learn discrete semantics between two sentences, we design a semantics-driven attention mechanism.
    Specifically, given the quantized representations $\vb*{Z}_i^{q}$, $\vb*{Z}_j^{q}$ for sentence $\vb*{X}_i$ and $\vb*{X}_j$ respectively, we 
    compute the quantized representation-based query and key matrices as follows:
    
    {\small
    \begin{align}
        \vb*{Q}_q& = \vb*{Z W}_q^{Q},~~~\vb*{K}_q = \vb*{Z W}_q^{K},
        ~~~\vb*{Z} = \text{FFN}[\vb*{Z}_i^{q} \oplus \vb*{Z}_j^{q}],
    \end{align}  
    }
    where $\vb*{Z}$, $\vb*{Q}_q$, $\vb*{K}_q \in \mathbb{R}^{L\times E}$, $\vb*{W}_q^{Q}$ and $\vb*{W}_q^{K} \in \mathbb{R}^{E\times E}$, FFN means a fully connected feed-forward network. 
    The semantics-driven  attention can be defined as:
    \begin{equation}\small
        \text{Attention}(\vb*{Q},\vb*{K},\vb*{V}) = \text{Softmax}(\dfrac{\vb*{QK}^{T}+\vb*{Q}_q\vb*{K}_q^{T}}{\sqrt{d_k}}) \vb*{V}.
        \label{eq-13}
    \end{equation}
    It is different from the topic select-attention block introduced by \cite{lu2021topicaware}, which is an attention block built separately on top of the multi-head attention block.
    In addition to the introduction of the quantized representations into the self-attention mechanism,
    we also leverage those representations to enhance the final output $\vb*{h}^l$ of multi-layer transformer, i.e., $[\vb*{h}^l;\vb*{Z}_i^{q} \oplus \vb*{Z}_j^{q}] \in \mathbb{R}^{L \times (n+1)E} $, which is further fed into a non-linear layer to predict label $Y$ for semantic textual matching.
    
    \subsection{Joint Training}
    Our proposed framework integrates the three modules in Figure \ref{fig:Framework}, i.e., the NTM, the topic-enhanced VQ-VAE, and the semantics-attended Transformer equipped with the semantics-driven attention mechanism.
    We first pretrain the NTM to obtain meaningful topic information used to initialize the codebook of the topic-enhanced VQ-VAE.
    Then we joint train the NTM and topic-enhanced VQ-VAE to boost their performance.
    The joint loss function is defined as:
        \begin{align}
            L = L_{VQ-VAE} + \gamma L_{NTM},
        \end{align}       
    where $\gamma$ is the trade-off parameter controlling the balance between the topic model and the topic-enhanced VQ-VAE model.
    Finally, we incorporate the learned quantized representations from topic-enhanced VQ-VAE into the transformer block and fine-tune the transformer block on the STS task.
    
    \label{joint training}

\section{Experiment and analysis}
    \subsection{Datasets}
        We conducted experiments on the following benchmark datasets 
        used for the STS task, which include MRPC, Quora, STS-B, and SemEval CQA. 
         The Microsoft Research Paraphrase dataset ({\bf MRPC}) ~\cite{MRPC} contains pairs of sentences from news websites with binary labels for paraphrase detection.
         The {\bf Quora} duplicate question dataset contains more than 400k question pairs with binary labels
         for predicting if two questions are paraphrases. 
         We use train/dev/test set partition from \cite{Quora}. 
        {\bf STS-B}  is a collection of sentence pairs extracted from news headlines and other sources. It comprises a selection of the English datasets used in the STS task which were annotated with a score from 1 to 5 denoting how similar the two sentences are.
         The {\bf SemEval} community question answering has three subtasks: {\bf (A)} Question–Comment Similarity, {\bf(B)} Question–Question Similarity, {\bf(C)} Question–External Comment Similarity~\cite{semeval-2015,semeval-2016,semeval-2017}.  Following their settings, we use the 2016 test set as the development set and the 2017 test set as the test set.
         Table \ref{datasets} summarises their statistics. 
          
        \begin{table}[!t]
        \centering
        \small
        \begin{tabular}{c|cccc}
        \hline
            Dataset&{Train/Dev/Test}&Len&Label\\
            \hline
                MRPC     &3,668/408/1,725 & 22 &0/1\\
                Quora       &384,348/10,000/10,000 &13 &0/1\\
                STS-B  &5,749/1,500 /1,379 &10   &0\textasciitilde 5\\
                SemEval-A &20,340/3,270/2,930 & 41&0/1\\
                SemEval-B &3,169/700/880 & 46 &0/1\\
                SemEval-C &31,690/7,000/8,800  & 39 &0/1\\
        \hline
        \end{tabular}
        \caption{Statistics of Datasets}
        \vspace{-6mm}
        \label{datasets}
        \end{table} 
      
        \begin{table*}[!ht] 
            \centering
            \resizebox{\columnwidth*2}{!}{%
            \begin{tabular}{lccccccccccccccc}
                \hline
                \multirow{2}*{Models} & \multicolumn{2}{c}{\multirow{2}*{MRPC}} & \multicolumn{2}{c}{\multirow{2}*{Quora}} &\multicolumn{2}{c}{\multirow{2}*{STS-B}} &\multicolumn{6}{c}{SemEval} \\ 
                \cmidrule(r){8-9} \cmidrule(r){10-11} \cmidrule(r){12-13}
                &&&&&&&\multicolumn{2}{c}{A}&\multicolumn{2}{c}{B}&\multicolumn{2}{c}{C}\\
                
                \hline 
                {\bf Baselines}& $Acc$ & $F1$  & $Acc$ & $F1$ & $PC$ & $SC$  & $Acc$ & $F1$ & $Acc$ & $F1$  & $Acc$ & $F1$ \\
                ERNIE   &83.97 &88.16 &90.88&90.76&85.12&83.26&77.53&75.10&68.86 &51.42 &94.36&27.20\\
                tBert &84.29 &88.44&90.76&90.65&85.34&83.96&78.02&76.77&71.44 &52.10&94.07&27.33 \\
                SemBert &84.02 &88.24&90.83&90.77&\bf87.13&85.34&78.15&76.33&71.47 &52.19 &94.60&27.44 \\
                
                Bert  &84.21 &88.21 &90.79&90.72&86.11&84.90&77.88&75.25&66.81 &51.33 &94.40&27.36   \\
                {\bf DisBert} &\bf{84.70}& \bf{89.06} &\bf90.94&\bf90.81&86.67&\bf85.64&\bf79.15&\bf77.57&\bf72.38 &\bf54.44 &\bf94.96&\bf27.61  \\
                \hline
                RoBERTa &87.01 &90.61&91.03&91.45&89.04&88.38&78.53&76.56&75.09&55.75 &93.52&32.46 \\
                {\bf DisRoBERTa}   &\bf88.06 &\bf91.15 &\bf91.53&\bf91.81&\bf89.90&\bf89.28&\bf78.89&\bf77.01&\bf77.72&\bf57.18 &\bf95.63&\bf33.79    \\
                \hline
            \end{tabular}
            }
            \vspace{-3mm}
            \caption{Comparisons of performance on six STS datasets. We report average performance for five different random seeds. The better results of different transformer-based model are highlighted in bold (according to the pairwise t-test with 95\% confidence).}  
        \label{overallPeromance}
        \vspace{-5mm}
        \end{table*}  
        
        \begin{figure*}
        \centering
        \subfigure[MRPC]{
            \begin{minipage}[t]{0.5\textwidth}
            \centering
            \includegraphics[width=1\textwidth]{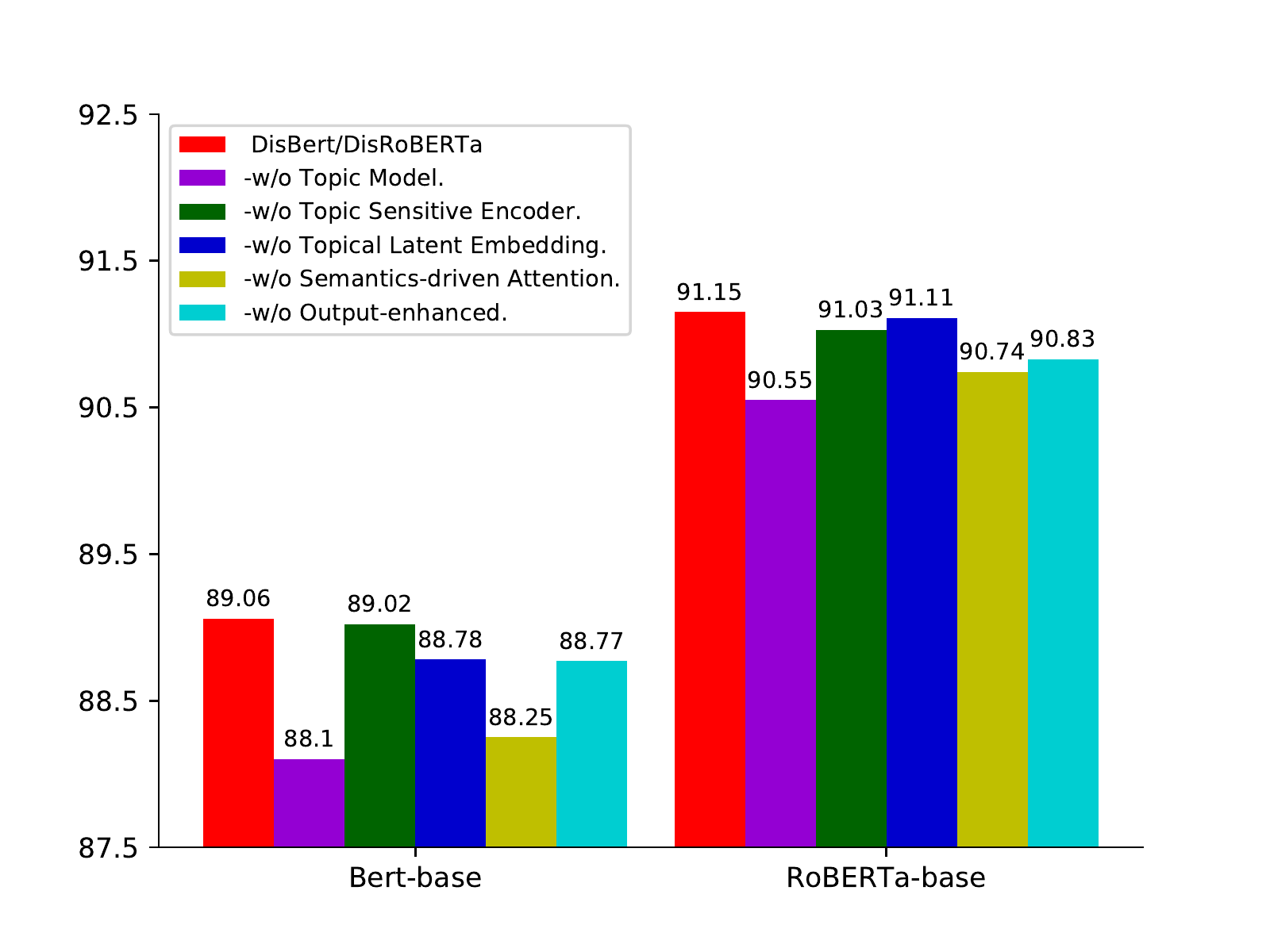}
            \vspace{-5mm}
            \end{minipage}%
            }%
        \subfigure[Quora]{
            \begin{minipage}[t]{0.5\textwidth}
            \centering
            \includegraphics[width=1\textwidth]{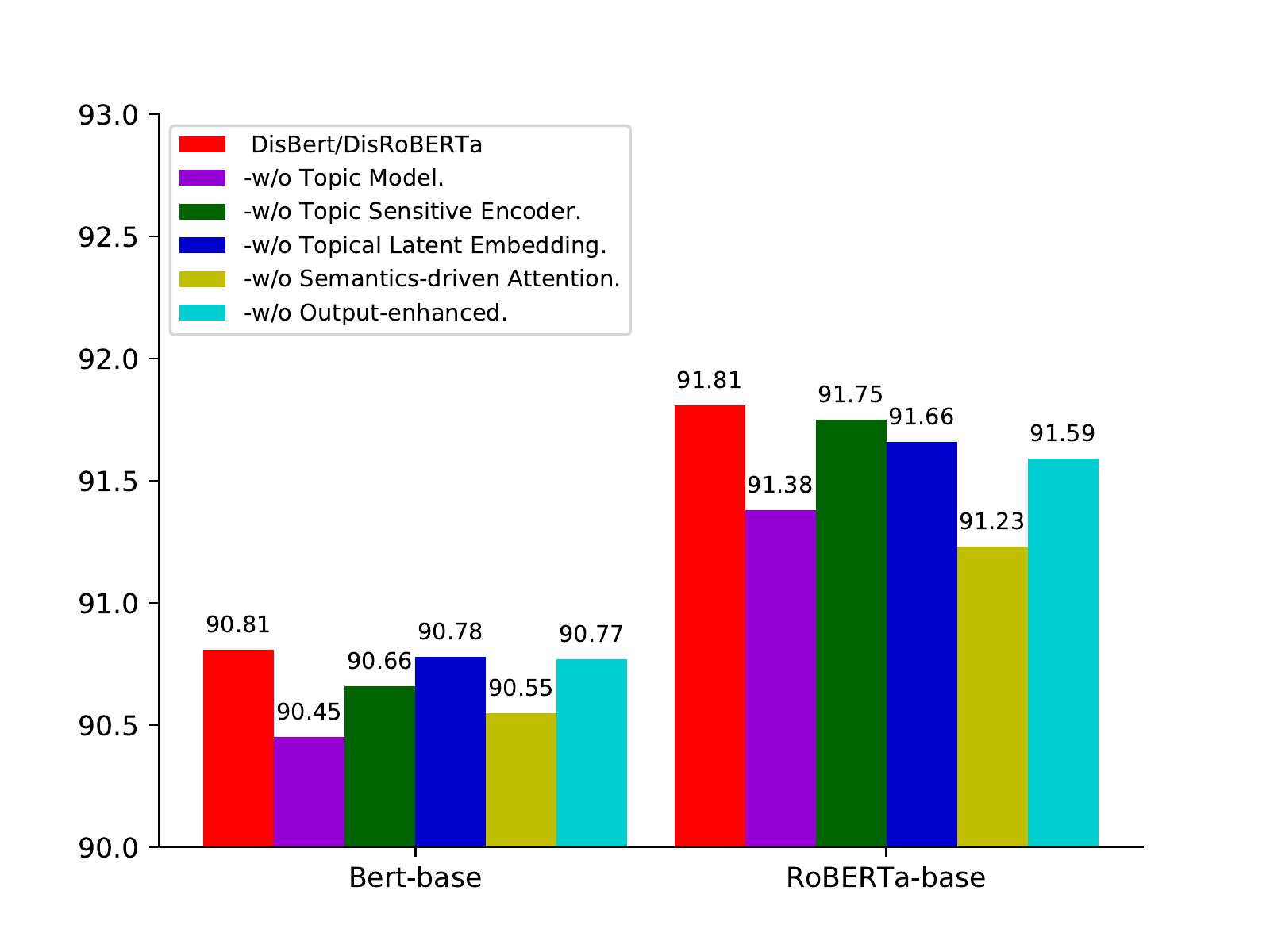}
            \vspace{-5mm}
        \end{minipage}%
        }%
        \centering
        \vspace{-3mm}
        \caption{Ablation study on DisBert and DisRoBERTa model. We reported F1 score for MRPC and Quora datasets.}
        \label{ablation}
        \vspace{-5mm}
        \end{figure*}    
    
    \subsection{Baseline Models}
    We implemented several strong baselines in the STS task for comparison.~To evaluate the semantics-attended Transformer, we select two widely-used pretrained language models, {\bf Bert}~\cite{BERT} and {\bf RoBERTa}~\cite{RoBERTa}. 
    We used the small version to fine-tune the STS task, i.e., Bert-base and RoBERTa-base.~And we modified these two models by adding the semantics-driven multi-head attention to the original multi-head attention,
    which is denoted as {\bf DisBert} and {\bf DisRoBERTa} respectively.
    We selected {\bf tBERT}~\cite{tbert} as an important baseline that also makes use of topic information. 
    Different from tBert which directly concatenates topic representation and sentence pair vector, we leverage the topic model to help VQ-VAE capture global semantics through the topic sensitive encoder and topical latent embedding.
    We also select the knowledge-enhanced Bert model {\bf ERNIE} ~\cite{ERNIE} that incorporates knowledge graph into Bert and achieves improvements on language understanding tasks.
    To further verify the performance on the STS task, we also compared with Semantic-aware Bert ({\bf SemBert}) ~\cite{semBert}, which incorporates explicit contextual semantics and outperforms other Bert-based models on the STS task.

    \subsection{Experimental Settings}
        For all methods, we performed a greedy search to find their optimal hyper-parameters using the development set. 
        For NTM, we processed the datasets with gensim tokenizer
        \footnote{https://radimrehurek.com/gensim/ utils.html}, and a vocabulary for each dataset was 
        built based on its training set with stop words and words occurring less than 3 times removed. 
        The BoW input of topic model  $\vb*{X}^{BoW}$ on each dataset was constructed based on the corresponding vocabulary.
        For the VQ-VAE model, we set the dimension of each code $E$ to 64 and commitment loss $\lambda$ = 0.0001. 
       \footnote{Code is available at https://github.com/ErxinYu/DisBert.}
        
        For joint training, we pretrained the NTM and selected the best model based on their reconstruction perplexity.
        Then we jointly trained the selected NTM and topic-enhanced VQ-VAE with $\gamma$ = 1.
        We fine-tuned our semantics-driven language model and the vanilla one with the same parameters for a fair comparison, following their settings with the publicly available code.
        We use  Pearson correlation coefficient (PC) and Spearman correlation coefficient (SC) for STS-B datasets. 
        For other datasets, we report accuracy (ACC) and F1 score.
        More detailed parameter settings are described in the appendix.

    \subsection{Performance on Six STS Datasets}
        Table \ref{overallPeromance} shows the results drove on the six 
        datasets, for which
        we have the following observations.
        \begin{itemize}[topsep=0pt, partopsep=0pt, noitemsep, leftmargin=*]
            \item  {\bf Semantics-attended transformer is effective.} Our model, DisBert, which introduces the quantized representations into the transformer, significantly outperforms Bert, with the improvement ranging from $0.1\% $ to $5.5\% $ on the six datasets. 
            The results demonstrate the usefulness of our semantics-driven multi-head attention mechanism.
            Similarly, RoBERTa equipped with vector quantized representations learned by our topic-enhanced VQ-VAE
            also achieves improved performance.
            Thus, the results on both Bert and RoBERTa verify that our semantics-driven attention is an effective plug-and-play module that can be readily applied to various transformer-based language models.
            \item  {\bf DisBert outperforms other bert-based models.} 
            Our proposed methods achieve better performance than tBert does, 
            which directly concatenates the output of the topic model and Bert.
            It indicates that injecting topic-enhanced VQ-VAE into Bert is a more efficient way to provide global semantics.
            The performance compared with ERNIE shows that our method is better at capturing sentence semantics than incorporating knowledge graphs into Bert. 
            Compared with semBert, which is capable of explicitly absorbing contextual semantics, 
            our model gains better performance on five out of six STS datasets.
            Overall, the performance gain on almost datasets demonstrates the effectiveness of our proposed framework.  
            
        \end{itemize}
    
            \begin{figure*}[!htb]
            \centering
            \subfigure[SemEval-A]{
                \begin{minipage}[t]{0.33\textwidth}
                \centering
                \includegraphics[width=1\textwidth]{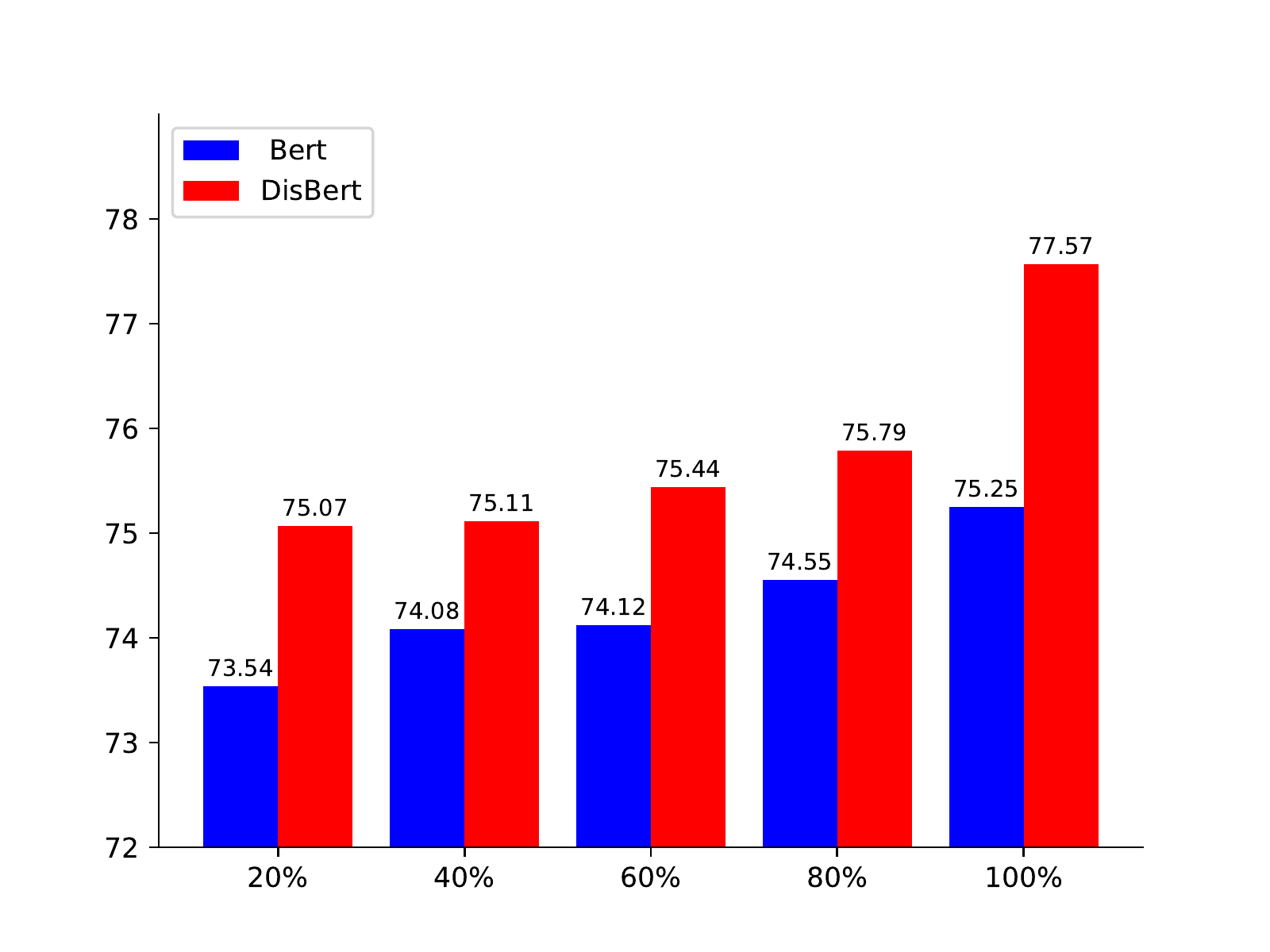}
                \end{minipage}%
            }%
            \subfigure[SemEval-B]{
                \begin{minipage}[t]{0.33\textwidth}
                \centering
                \includegraphics[width=1\textwidth]{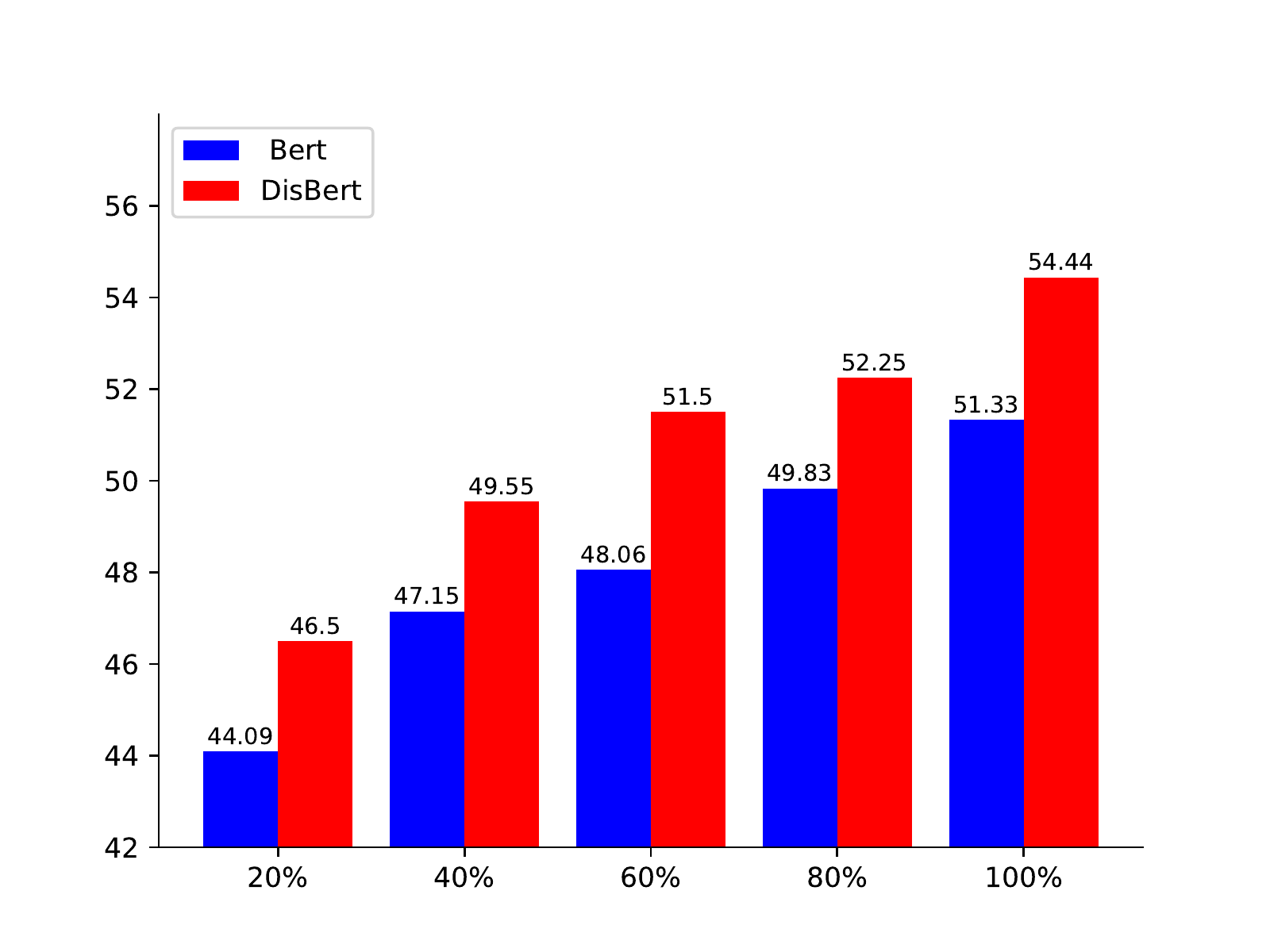}
                \end{minipage}%
            }%
            \subfigure[SemEval-C]{
                \begin{minipage}[t]{0.33\textwidth}
                \centering
                \includegraphics[width=1\textwidth]{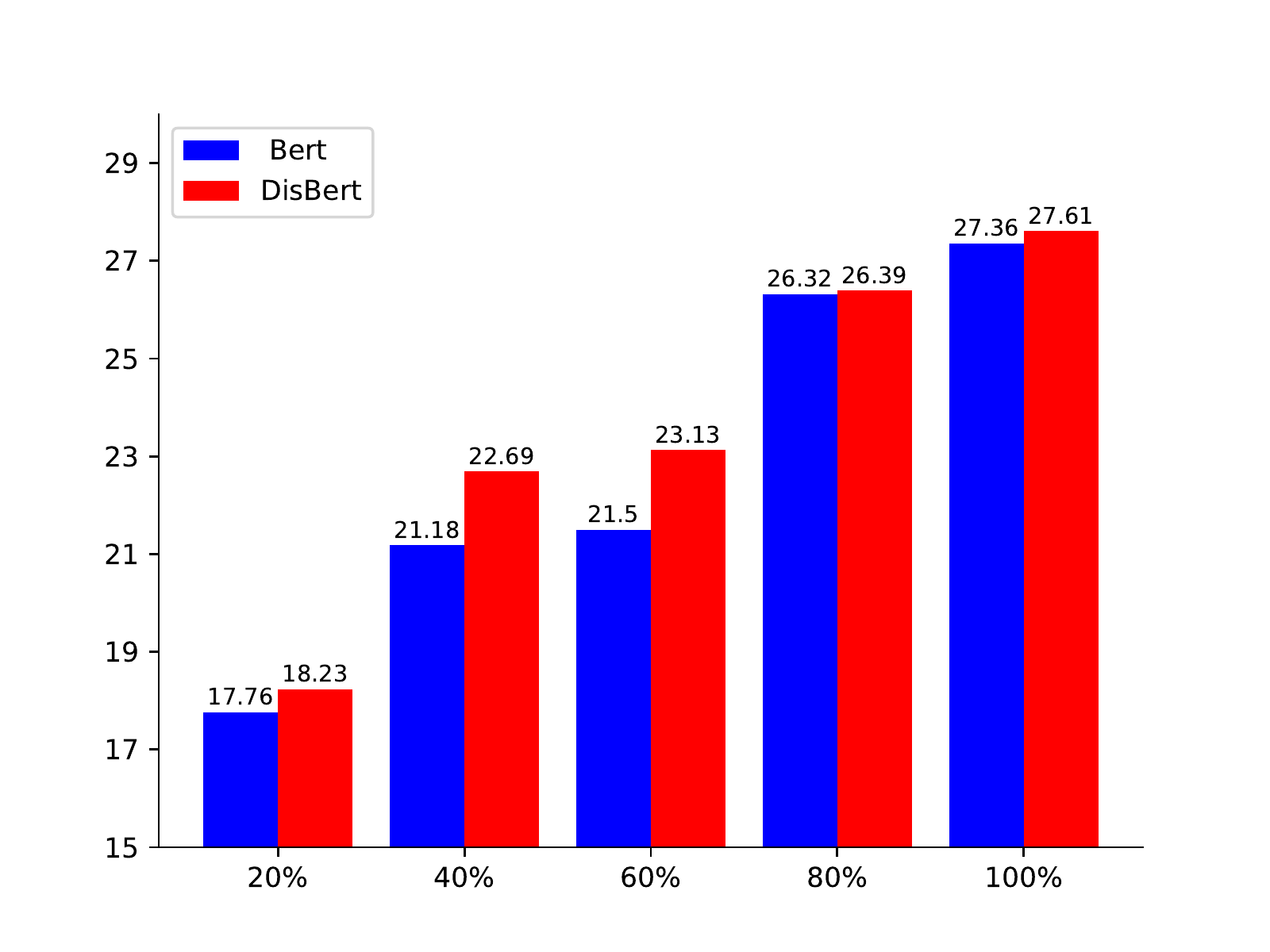}
                \end{minipage}%
            }%
            \vspace{-3mm}
            \caption{Performance of Bert and DisBert with different amounts of training data of SemEval datasets.}
            \label{data_amount}
            \vspace{-3mm}
        \end{figure*}

        \begin{figure}[htb]
            \centering
            \subfigure[MPRC]{
                \begin{minipage}[t]{0.25\textwidth}
                \centering
                \includegraphics[width=1\textwidth]{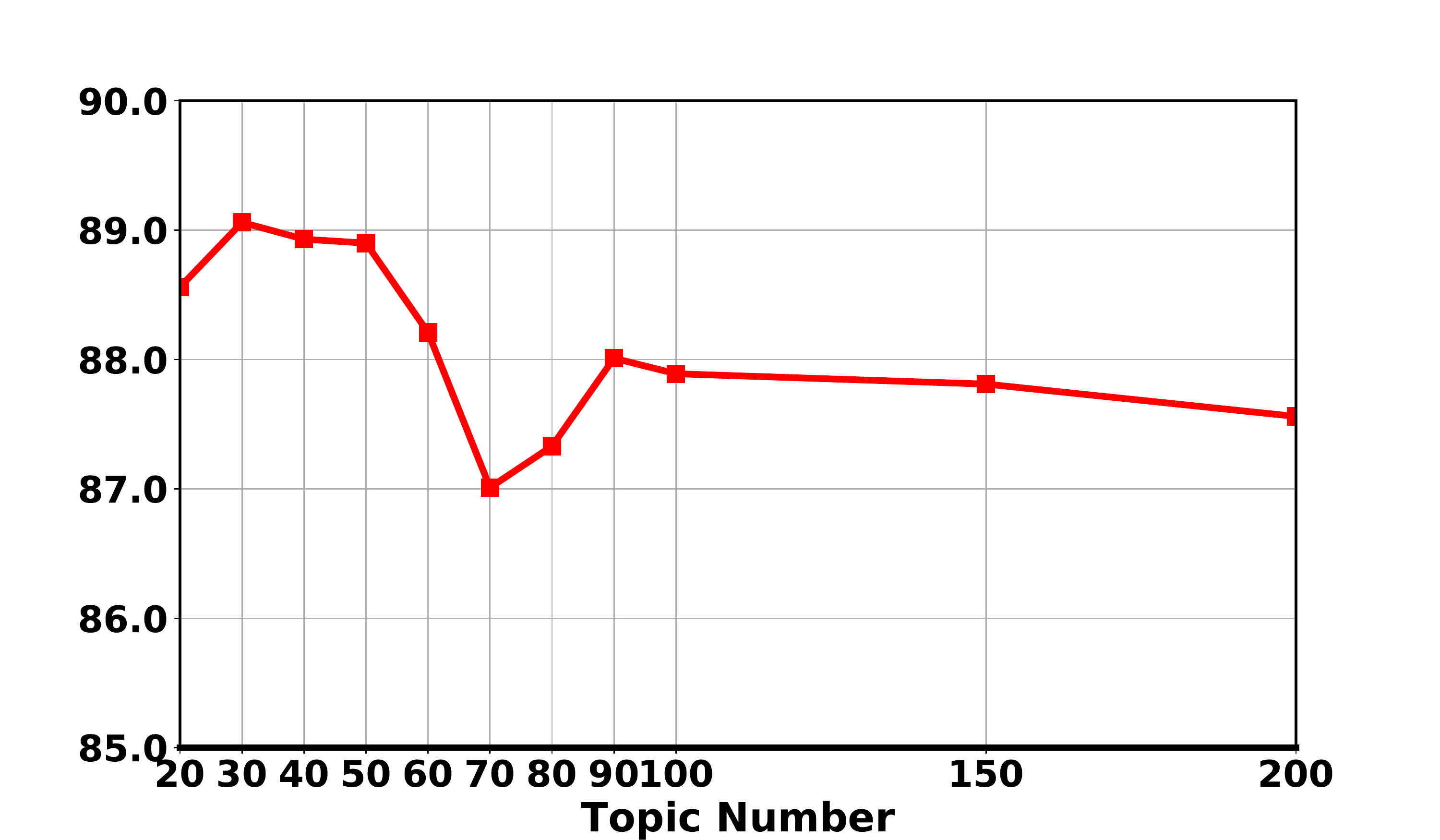}
            \end{minipage}%
            }%
            \subfigure[Quora]{
                \begin{minipage}[t]{0.25\textwidth}
                \centering
                \includegraphics[width=1\textwidth]{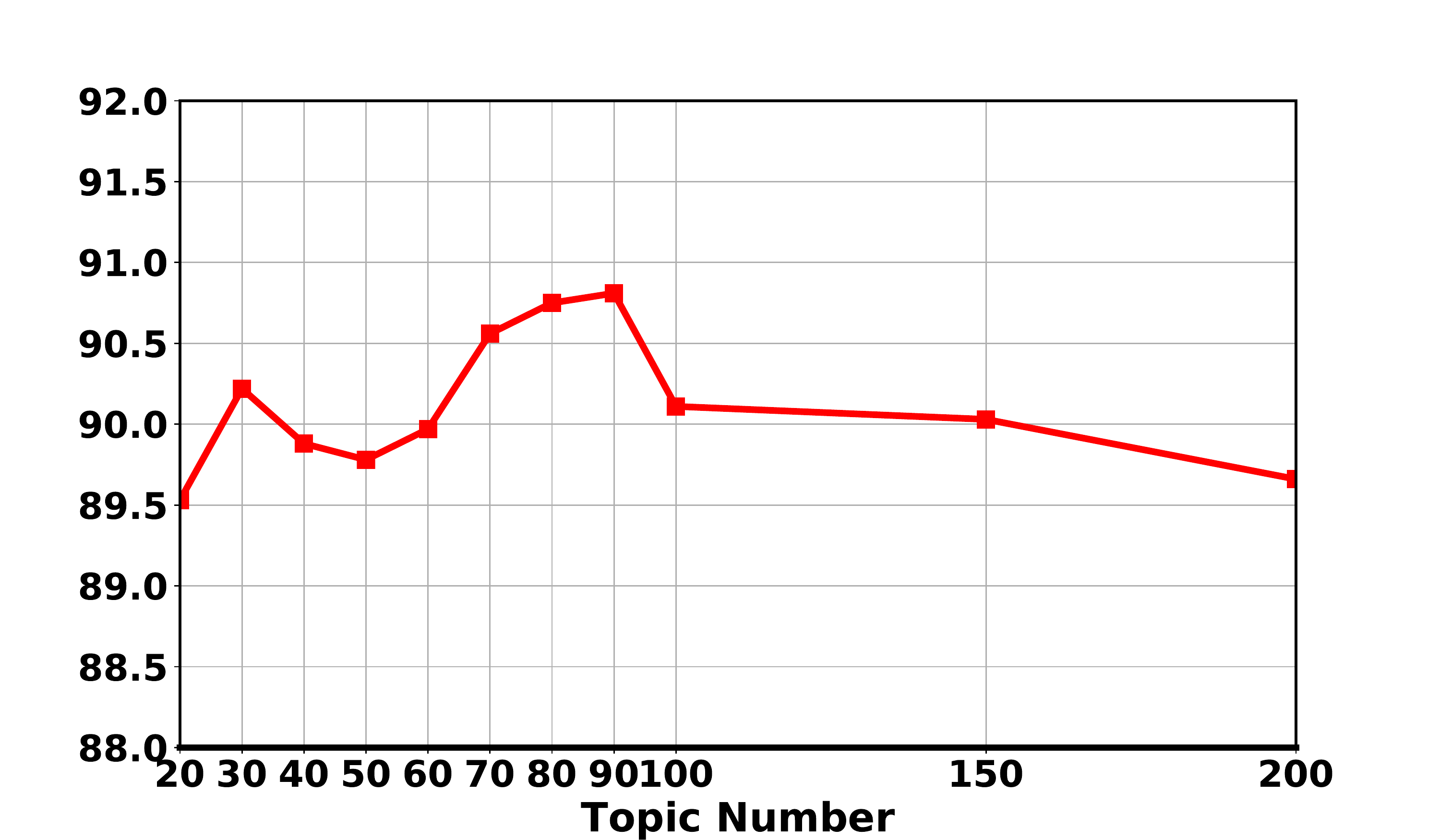}
            \end{minipage}%
            }%
            
            \subfigure[STS-B]{
                \begin{minipage}[t]{0.25\textwidth}
                \centering
                \includegraphics[width=1\textwidth]{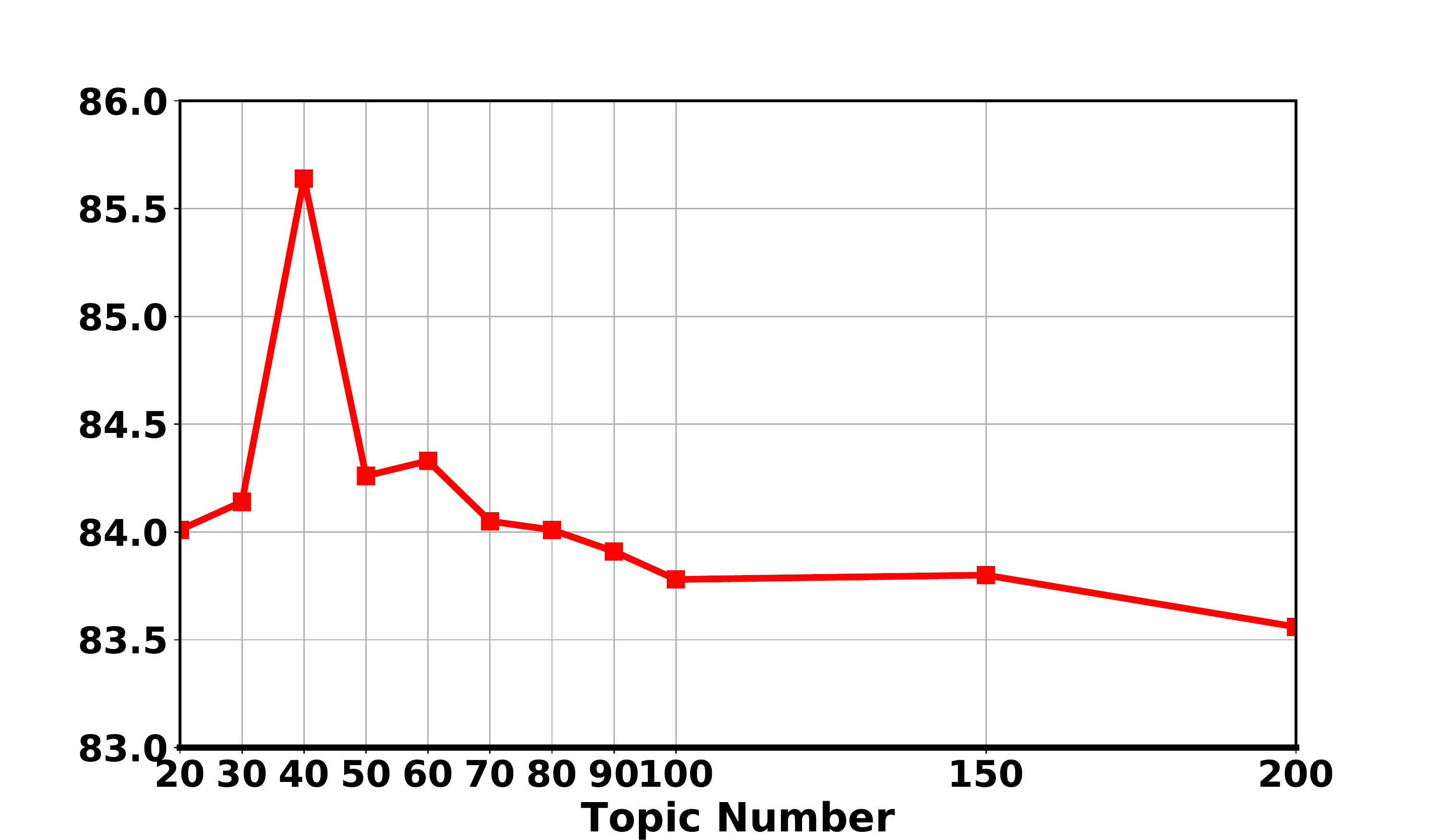}
            \end{minipage}%
            }%
            \subfigure[SemEval-A]{
                \begin{minipage}[t]{0.25\textwidth}
                \centering
                \includegraphics[width=1\textwidth]{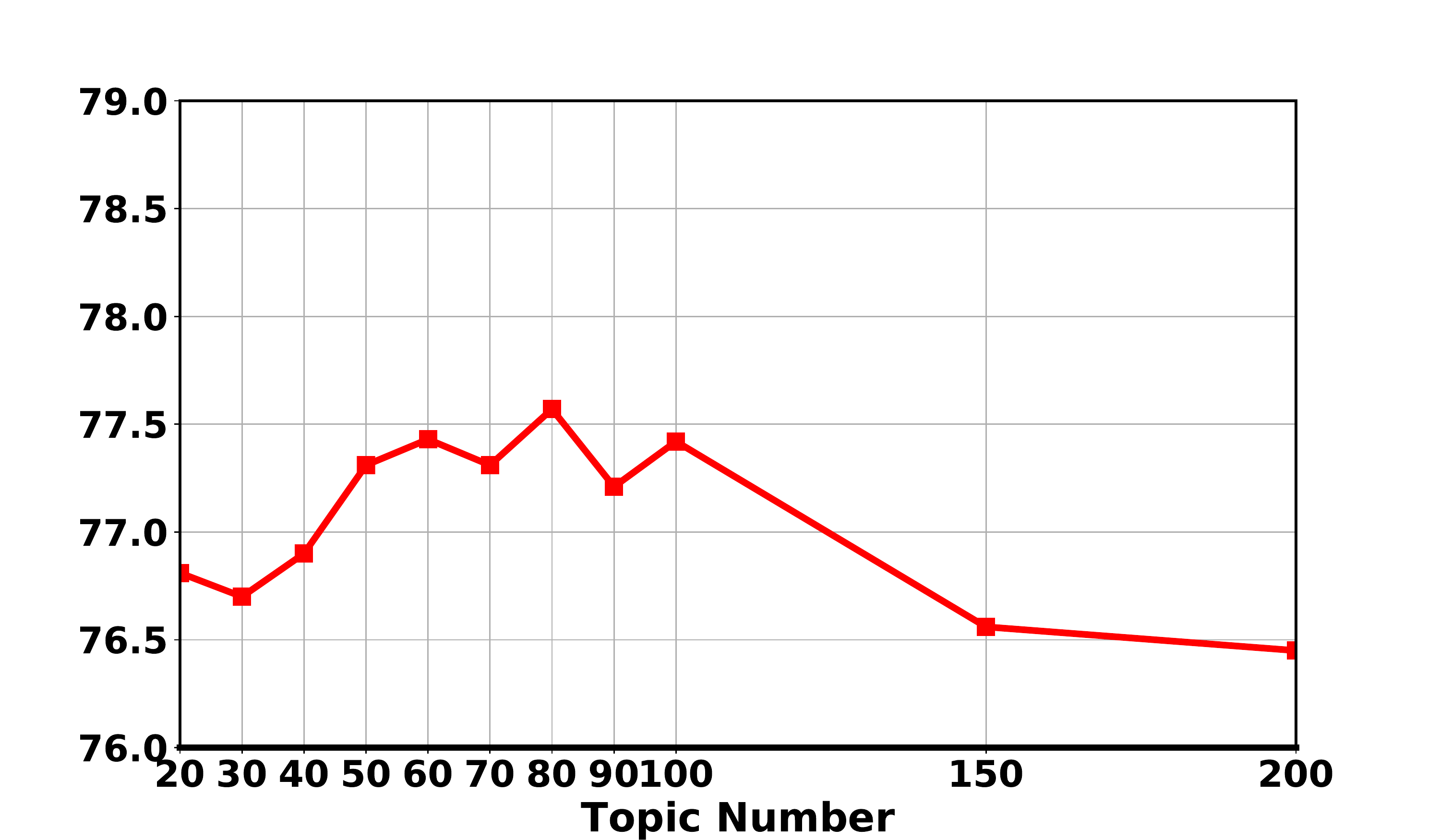}
            \end{minipage}%
            }%
            
            \subfigure[SemEval-B]{
                \begin{minipage}[t]{0.25\textwidth}
                \centering
                \includegraphics[width=1\textwidth]{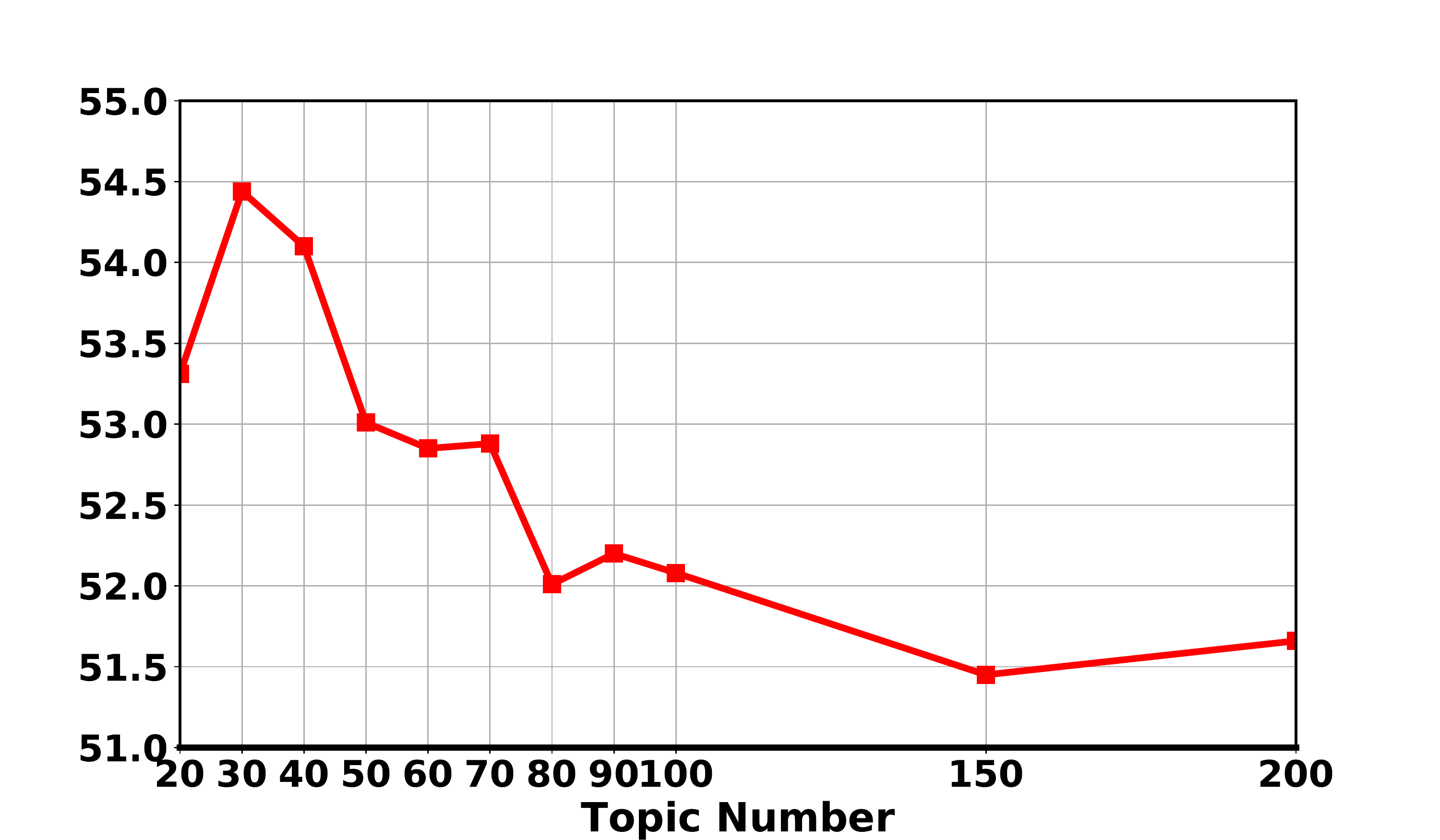}
            \end{minipage}%
            }%
            \subfigure[SemEval-C]{
                \begin{minipage}[t]{0.25\textwidth}
                \centering
                \includegraphics[width=1\textwidth]{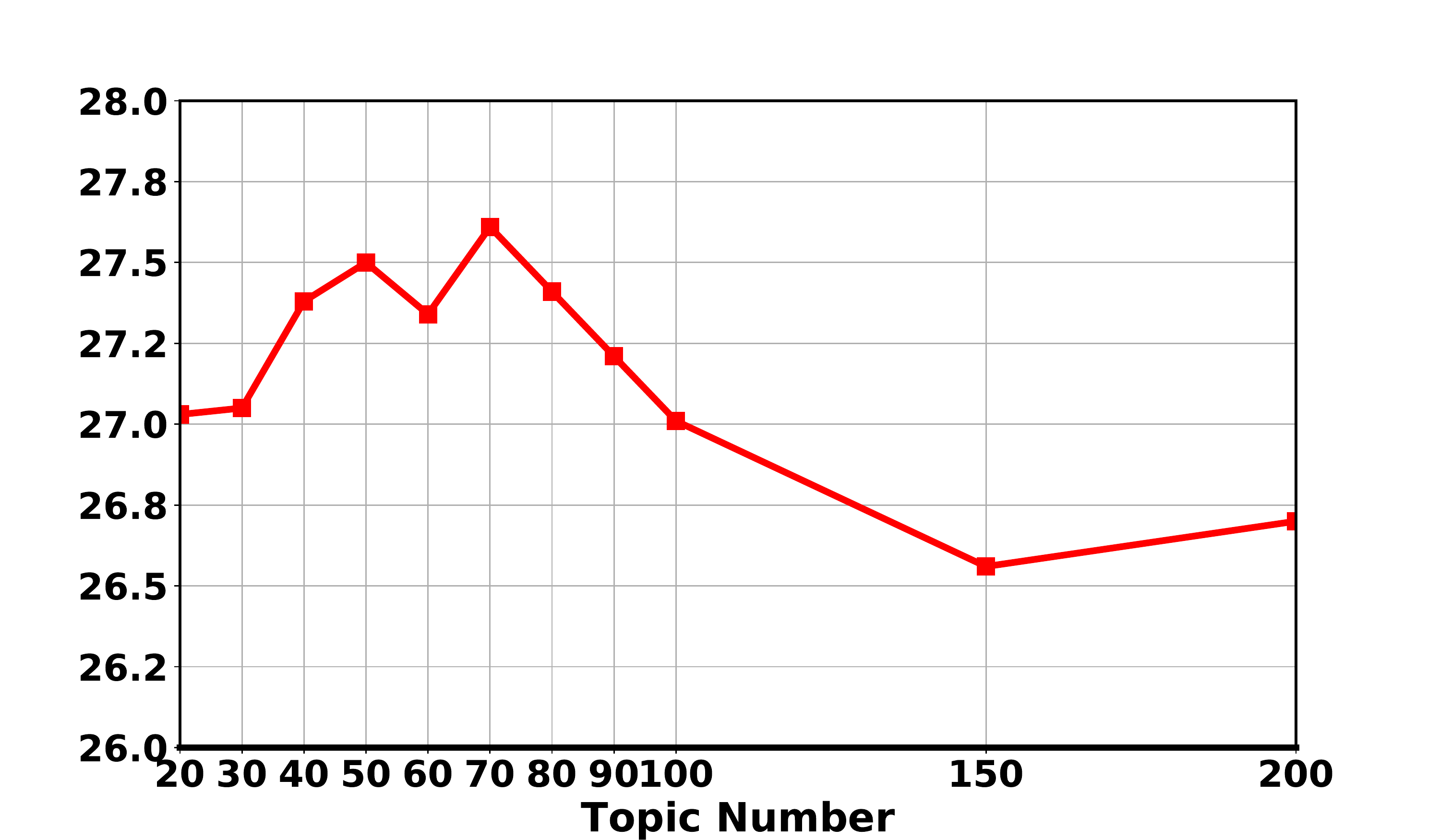}
            \end{minipage}%
            }%
        \centering
        \vspace{-3mm}
        \caption{The impact of topic number $K$ on DisBert, where the horizontal axis shows the number of topics and the vertical axis shows the performance on STS datasets.
        Spearman correlation coefficient is reported in STS-B datasets and F1 scores are reported for other datasets.}
        \label{k_impact}
        \vspace{-6mm}
        \end{figure}

    \subsection{Ablation Study}
        We compared different variants of DisBert and DisRoBERTa in order to study how topic-informed discrete latent variables
        affect performance. 
        {\bf w/o Topic Model} means using the vanilla VQ-VAE model to get discrete latent variables without topic information.
        {\bf w/o Topic Sensitive Encoder} refers to removing topical embedding $\vb*{X^t}$ in the encoding phase  and {\bf w/o Topical Latent Embedding} means randomly initializing embedding instead of the topic  embedding $\vb*{\beta}$.
        {\bf w/o Semantics-driven Attention} stands for the variant excludes the attention scores calculated by discrete variables (i.e., $\vb*{Q_qK^t_q}$) in Eq~\eqref{eq-13}.
        {\bf w/o Output-enhanced} is the variant does not concatenate quantized variable $\vb*{Z^{q}}$ with $\vb*{h}^l$ in the output layer.
        
        As shown in Figure \ref{ablation} we have the following observation: 
        (1) Topic information is informative for the STS task for both DisBert and DisRoBERTa.
        The performance of DisBert has dropped 0.96, 0.04, and 0.28 on MRPC datasets without the topic model, topic sensitive encoder, and topical latent embedding respectively.
        (2) The well-designed semantics-driven multi-head attention benefits the STS task. 
        Without such attention, performance dropped for both DisBert and DisRoBERTa on both datasets. Meanwhile, when the output is equipped with quantized representations, performance is significantly improved, which further verifies that the learned topic-informed discrete latent variable can capture semantics and improve the STS~task.
        
        \begin{table}[!]
            \centering
            \resizebox{\columnwidth}{!}{
                \begin{tabular}{ccccc}
                    \hline
                      &\multicolumn{2}{c}{DisBert}&\multicolumn{2}{c}{DisRoBERTa}\\
                       
                    \cmidrule(r){2-3} \cmidrule(r){4-5}
                    &Multistage&Joint&Multistage&Joint\\
                    \hline
                    {\bf MRPC} &88.78 &\bf89.06 &91.04 &\bf91.15\\
                    {\bf Quora}    & 90.45&\bf90.81&90.67&\bf91.81\\
                    {\bf STS-B}    & 85.32&\bf85.64&88.89&\bf89.28\\
                    {\bf SemEval-A}    & 77.49&\bf77.57&76.89&\bf77.01\\
                    {\bf SemEval-B}    & 54.21&\bf54.44&56.95&\bf57.18\\
                    {\bf SemEval-C}    & 27.50&\bf27.61&33.57&\bf33.79\\
                    \hline
                \end{tabular}
                }
                \vspace{-3mm}
            \caption{Joint training evaluation on six datasets. Spearman correlation coefficient is reported in STS-B datasets and F1 scores are reported for other datasets.}
            \label{joint}
            \vspace{-5mm}
            \end{table}

            \begin{figure*}[!htb]
                \centering
                \vspace{-11mm}
                \subfigure[Bert-based]{
                    \begin{minipage}[t]{0.5\textwidth}
                    \centering
                    \includegraphics[width=1\textwidth]{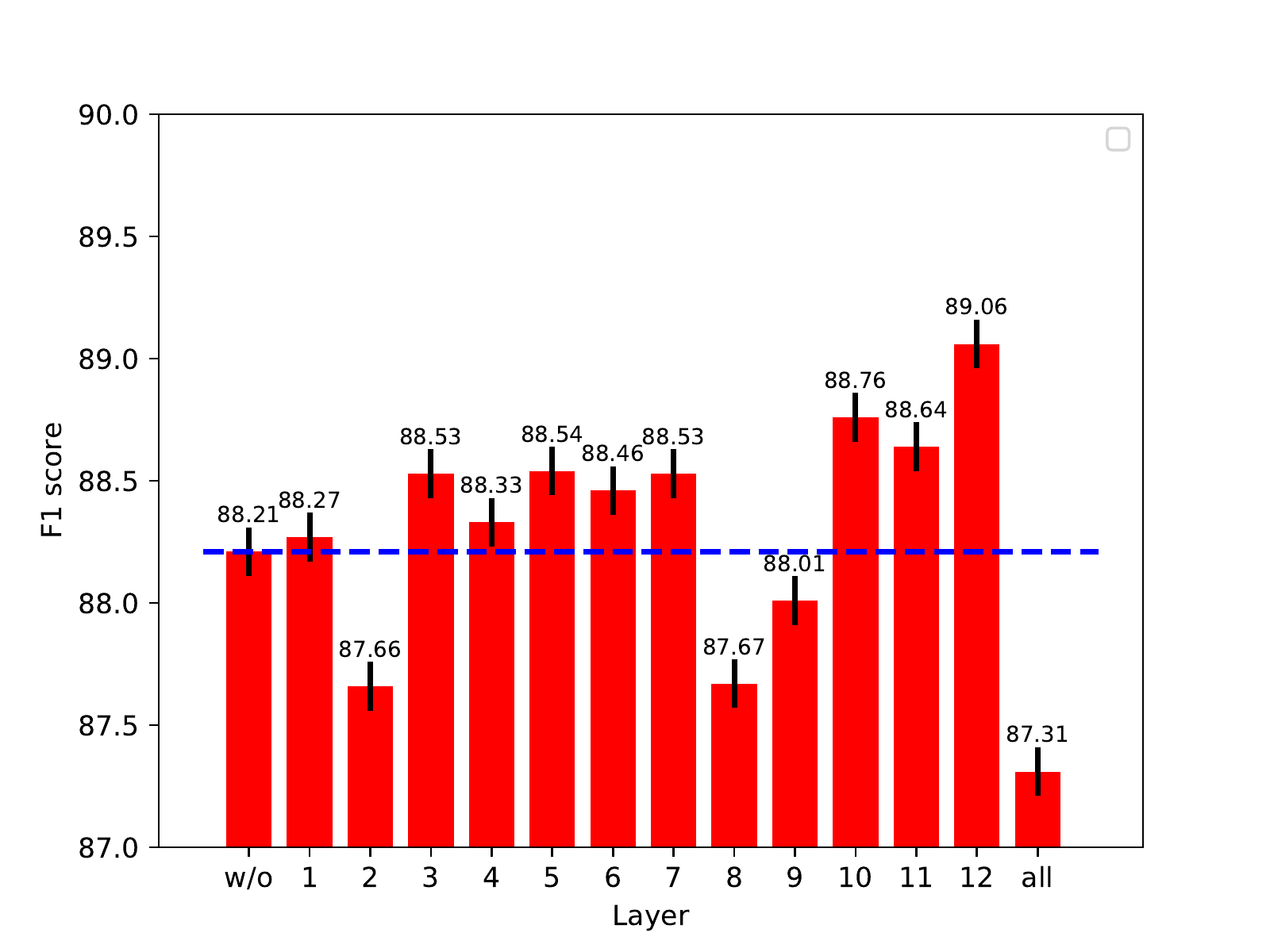}
                    \end{minipage}%
                    }\subfigure[RoBERTa-based]{
                    \begin{minipage}[t]{0.5\textwidth}
                    \centering
                    \includegraphics[width=1\textwidth]{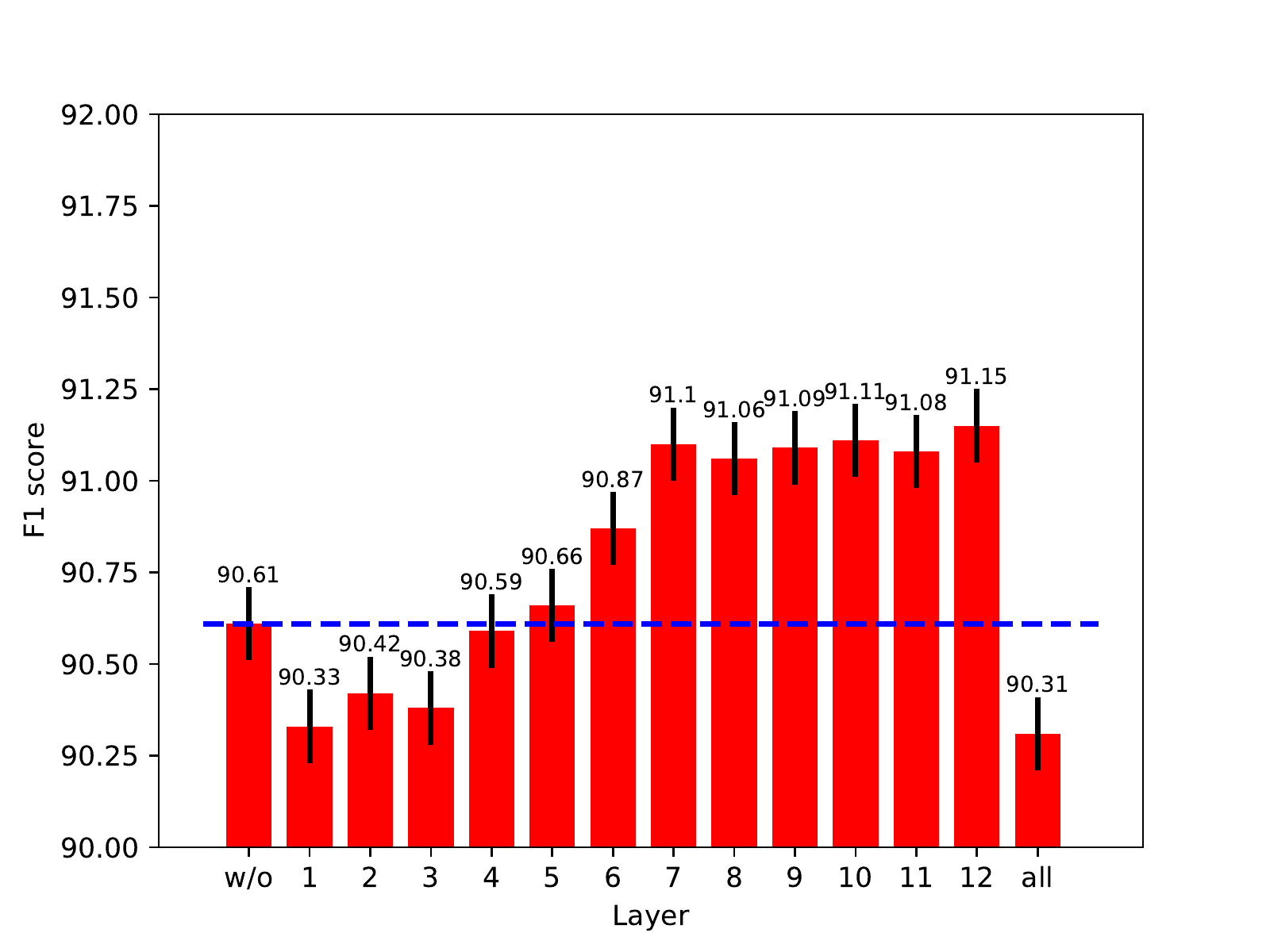}
                    \end{minipage}%
                    }%
                \centering
                \caption{Performance of DisBert and DisRoBERTa with semantics-driven multi-head attention on different layers. {\bf w/o} means the origin Bert/RoBERTa without aforesaid attention and {\bf all} means such attention is applied to all layers.}
                \label{layer_impact}
                \vspace{-2mm}
                \end{figure*}

                \begin{table*}[htb]
                    \centering
                    \begin{subtable}
                           \centering
                          \resizebox{0.85\columnwidth}{!}{
                            \begin{tabular}{ll}
                            \toprule
                            Cluster &Words \cr
                            \midrule
                            7  &dead  protesters attack crash  weapons kill killed\\ 
                            16  &bus day transport by car  bike has what  \\ 
                            20 &government vote year elections obama winner \\
                            26 &in for you on next out  it up before us with  \\
                            \bottomrule
                            \end{tabular}
                            }
                            \label{cluster words}
                      \end{subtable}\begin{subtable}
                          \centering
                          \resizebox{1.15\columnwidth}{!}{
                            \begin{tabular}{ll}
                            \toprule
                                Sentence Pair&Label \\
                            \hline
                                S1: 
                                ${\textbf {\rm \color{black}{Five}} }$ 
                                ${\textbf {\rm \color{red}{killed}}_{\color{red}{[7]}}}$ 
                                ${\textbf {\rm \color{blue}{in}}_{\color{blue}{[26]}}}$
                                ${\textbf {\rm \color{black}{Belgian}}}$
                                ${\textbf {\rm \color{green}{coach}}_{\color{green}{[16]}}}$
                                ${\textbf {\rm \color{red}{crash}}_{\color{red}{[7]}}}$.& \multirow{2}*{3.8}\\ 
                                S2:  ${\textbf {\rm \color{black}{Teenagers}}}$
                                ${\textbf {\rm \color{blue}{among}}_{\color{blue}{[26]}}}$
                                ${\textbf {\rm \color{black}{5}}}$
                                ${\textbf {\rm \color{red}{dead}}_{\color{red}{[7]}}}$  
                                ${\textbf {\rm \color{blue}{in}}_{\color{blue}{[26]}}}$
                                ${\textbf {\rm \color{black}{Belgian}}}$
                                ${\textbf {\rm \color{green}{bus}}_{\color{green}{[16]}}}$
                                ${\textbf {\rm \color{red}{crash}}_{\color{red}{[7]}}}$ . &\\
                            \hline
                            S1: ${\textbf {\rm \color{black}{Mugabe}}}$
                              ${\textbf {\rm \color{red}{declared}}_{\color{red}{[20]}}}$
                            ${\textbf {\rm \color{red}{winner}}_{\color{red}{[20]}}}$
                            ${\textbf {\rm \color{black}{of}}}$
                            ${\textbf {\rm \color{black}{disputed}}}$
                            ${\textbf {\rm \color{red}{elections}}_{\color{red}{[20]}}}$& \multirow{2}*{4.0} \\ 
                            S2: ${\textbf {\rm \color{black}{Zimbabwe}}}$
                            ${\textbf {\rm \color{black}{Mugabe}}}$
                            ${\textbf {\rm \color{red}{declared}}_{\color{red}{[20]}}}$
                            ${\textbf {\rm \color{red}{winner}}_{\color{red}{[20]}}}$
                            ${\textbf {\rm \color{black}{in}}}$ 
                            ${\textbf {\rm \color{black}{disputed}}}$
                            ${\textbf {\rm \color{red}{vote}}_{\color{red}{[20]}}}$\\         
                            \bottomrule
                            \end{tabular}
                            }
                            \vspace{-3mm}
                      \end{subtable}
                  \caption{Case Study on STS-B datasets. }
                  \vspace{-2mm}
                  \label{case_study}
                  \end{table*}    

    \subsection{Joint Training vs Multistage Training}
        After we pretrained the NTM, there are two ways of training our topic-enhanced VQ-VAE.
        One is to utilize the topic embeddings from the pretrained NTM as the codebook initialization in VQ-VAE and train the latter while holding the former fixed (denoted by \textbf{Multistage} in Table~\ref{joint}), the other is to joint train the pretrained NTM and VQ-VAE (denoted by \textbf{Joint}). 
        The results of these two training methods derived from the six STS datasets are shown in Table~\ref{joint}.
        We can see that compared with multistage training, 
        joint training yields better results on all datasets for both DisBert and DisRoBERTa.
        The comparison shows that jointly training the topic embedding matrix $\vb*\phi$ benefits the STS task.

    \subsection{Results with Varying Hyperparameters.} 
        \textbf{Impact of Training Data Amounts.} In figure \ref{data_amount}, we compared the results of Bert and DisBert with different training data.
        We randomly selected 20\% to 100\% data from the training set as training data.
        We can observe that DisBert outperforms Bert consistently across all training data sizes, which suggests that we can leverage the topic-enhanced discrete variable in all data sizes, even when the training data is scarce.
        
        \textbf{Impact of Topic Numbers.} Figure \ref{k_impact} shows the performance of DisBert given varying topic numbers. As we can see, the curves on all datasets are not monotonic and the best accuracy is achieved with different numbers of topics, e.g., $k=30$ on the MRPC dataset. 
        We also found that on larger training datasets DisBert requires larger topic numbers and vice versa, e.g., $k=90$ for Quora and $k=30$ for SemEval-B.
        It is not unexpected, as larger datasets can cover more topics.
        However, how to let the data choose the right number of topics in NTM is beyond the scope of this paper.
        
        \textbf{Impact of Bert Layers.} Figure \ref{layer_impact} shows the effect of semantics-driven multi-head attention on different layers. We found that the aforesaid attention is not always helpful for all layers and best performances are achieved when applying such attention to the last layer of DisBert/DisRoBERTa on MRPC datasets. 
        Meanwhile, the performance drops when such attention is applied to all layers.

             \begin{table*}[!htb]
            \centering
            \small
            \resizebox{0.8\textwidth}{!}{
            \begin{tabular}{c|cccc|c}
            \hline
                    Dataset& Topic Model &VQ-VAE & Semantics-driven Bert &Total&Bert\\
                    \hline
                        MRPC  &247 &218 &220 &685&200  \\
                        Quora   &3,512 &5,123 &18,232 &26,927&17,805  \\
                        STS-B  &197 &186 &239 &622&214  \\
                \hline
                \end{tabular}
            }
            \caption{Limitation---Training time (in seconds) of different models in three datasets. 
            We trained the topic model for 500 epochs, VQ-VAE for 20 epochs, semantics-driven Bert, and Bert for 10 epochs. All models are trained in the same environment with a single GeForce RTX 3090 card.}
            \vspace{-3mm} 
            \label{times}
            \end{table*}  
            
    \subsection{Case Studies}
        In Table~\ref{case_study}, we conduct a case study on STS-B datasets to show how the model works with topical information.
        The left part of the table shows four word clusters generated by Topic-enhanced VQ-VAE.
        As described in Section~\ref{VQVAE}, 
        the topic-enhanced VQ-VAE model clusters words in the discrete space given by the codebook.
        It can be seen that words in the same cluster form a meaningful topic:
        Cluster 7 is about protest, Cluster 6 is about transportation,
        Cluster 20 is about  government election, and Cluster 26 is one of the clusters capturing all the functional words.
        The right part of the table shows two pairs of sentences used in the STS tasks with the assignments of 
        the four topics to their words.
        We can observe that sentences, which are predicted to be similar by our model, tend to have more words assigned to similar topics, even if those words are morphologically different. 
        For instance, ``Killed'' and ``dead'' are both from Cluster 7, 
        ``coach'' and ``bus'' are from Cluster 16,
        and ``elections'' and ``vote'' are from Cluster 20.
        The heat map calculated by semantics-driven multi-head attention is shown on Appendix ~\ref{heatmap} for more discussion.

\section{Conclusion}
        In this paper,
        we developed a topic-enhanced VQ-VAE model to effectively train discrete latent variables by informing the vector quantization with semantics (i.e., latent topics) learned from a broader context by a neural topic model. 
        Then we further designed a semantics-driven multi-head attention mechanism for enriching the contextual embeddings learned by Transformer with topical information, which is calculated based on the  quantized representations from topic-enhanced VQ-VAE.
        Its plug-and-play characteristic allows it to be readily incorporated into various transformed-based language models.
        Through experiments on different scale datasets for the STS task, 
        we proved that building the semantic information learned by NTM via vector quantization into the multi-head attention block can improve the STS performance of transformer-based language models.


\section{Limitations}
    The limitations of this work, to the best of our knowledge, can be summarized into two aspects:
    
    (1) Compared to the origin Bert/RoBERTa, our model DisBert/DisRoBERTa needs a longer time to train since there are three components in our framework, i.e., topic model, VQ-VAE model, and the transformer-based model.
    From table ~\ref{times}, we can observe that in a small dataset like MRPC and STS-B, the training time is almost three times longer. 
    While in the large datasets Quora, our model took an extra 2.5 hours.
    Therefore in future work, we need to improve the efficiency of our model.

    (2) The improvement of our method is limited when sentences are long enough or training datasets are large, e.g., Quora and SemEvalC  shown in table \ref{overallPeromance}.
    Our model is capable of complete global semantics, hence it works better for small datasets and short sentences which contain limited semantics.

\section*{Acknowledgments}
This work is supported by the National Natural Science Foundation of China (No.61976102 and No.U19A2065) and the Science and Technology Development Program of Jilin Province (No.20210508060RQ), and the Fundamental Research Funds for the Central Universities, JLU.

\bibliography{anthology,custom}
\bibliographystyle{acl_natbib}

\appendix
\newpage
\section{Example Appendix}
\label{sec:appendix}
\subsection{Dataset Examples}
    Table ~\ref{datasets examples} shows examples from different datasets. 
    Labels indicate if the second sentence is a paraphrase (for paraphrasing tasks) or relevant (for QA tasks).
    In STS-B datasets, the label stands for sentence pair similarity score, and the higher score, the more similar the two sentences are.
        \begin{table*}[!t]
        \centering
        \small
        \resizebox{1\textwidth}{!}{%
        \begin{tabular}{clc}
        \hline
            Dataset&Sentence Pair&Label\\
            \hline
                \multirow{4}*{MRPC} &S1:The world's two largest automakers said their U.S. sales declined more than predicted last month  & \multirow{4}*{1}  \\
                &\quad \ \ \ as late summer sales frenzy caused more of an industry backlash than expected.\\
                &S2: Domestic sales at both GM and No. 2 Ford Motor Co. declined more than predicted as a late  &\\
                &\quad \ \ \ summer sales frenzy prompted a larger-than-expected industry backlash. &\\
                \hline
                \multirow{3}*{Quora} &S1: Currently , all Supreme Court Justices come from very elite law schools, is it similar for the best  & \multirow{3}*{1}  \\
                &\quad \ \ \ lawyers in private practice .\\
                &\multirow{1}*{S2:} What 's your type of jungle -LRB- concrete or nature -RRB- and why ? &\\
                \hline
                \multirow{2}*{STS-B} &S1: A man is spreading shreded cheese on a pizza.  & \multirow{2}*{3.8}  \\
                &\multirow{1}*{S2:} A man is spreading shredded cheese on an uncooked pizza. &\\
                \hline
                \multirow{2}*{SemEval-A} &S1: Massage oil is there any place i can find scented massage oils in qatar? & \multirow{2}*{0}  \\
                &\multirow{1}*{S2:} Whats the name of the shop?&\\
                \hline
                \multirow{3}*{SemEval-B} &S1: Music tastes so; what kind of music do you like? & \multirow{3}*{0}  \\
                &S2: After school program and summer camp Can anyone please tell me if there is any after school program \\
                &\quad \ \ \ and summer camp in Doha? -- for a grade two girl (seven years old).&\\
                \hline
                \multirow{3}*{SemEval-C} &S1: Best sunglass store in Qatar Can somebody suggest the best store where i can & \multirow{3}*{1} \\
                &\quad \ \ \ get the best sunglasses for ladies. Where can i find a store with the best variety?  \\
                &S2: Fashion wear - Primark Lingerie - Agent Provocateur Lingerie - Intimo Lingerie - Peach John \\
        \hline
        \end{tabular}
        }
        \caption{Dataset Examples.}
        \vspace{-3mm}
        \label{datasets examples}
        \end{table*} 
        
\subsection{Hyper-Parameter settings}
    Table \ref{parameter settings} shows the hyperparameter we chose in our model.
    All hyper-parameters were chosen through greedy search based on development set performance.
    
    For topic model, we choose number of topics in (20, 30, 40, 50, 60, 70, 80, 90, 100, 150, 200), batch size in (128, 256, 512), and learning rate in (1e-3, 2e-3, 3e-3).
    
    For the VQ-VAE model, we choose the batch size in (32, 64, 128), and the learning rate in (2e-5, 3e-5).
    
    For the Bert model, We choose the batch size in (32, 64, 128), and the learning rate in (2e-5, 3e-5).
         \begin{table*}[!ht] 
            \centering
            \resizebox{0.8\textwidth}{!}{%
            \begin{tabular}{lccccccccc}
                \hline
                \multirow{2}*{Models} & \multicolumn{1}{c}{\multirow{2}*{MRPC}} & \multicolumn{1}{c}{\multirow{2}*{Quora}} &\multicolumn{1}{c}{\multirow{2}*{STS-B}} &\multicolumn{3}{c}{SemEval} \\ 
                \cmidrule{5-7}
                &&&&\multicolumn{1}{c}{A}&\multicolumn{1}{c}{B}&\multicolumn{1}{c}{C}\\
                
                \hline 
                {\bf Topic Model} \\
                Topic number $K$ &30&90&40&80&30&70 \\
                Vocabulary size $V$ &5,179&37,013&4,369&12,102&3,763&13,272  \\
                Topic Embedding hidden size $E$ &64&64&64&64&64&64  \\
                Batch size &256&512&256&256&256&256& \\
                Epochs &500&150&500&200&500&200& \\
                
                Learning rate &1e-3&2e-3&1e-3&1e-3&1e-3&1e-3 \\
                \hline 
                {\bf VQ-VAE Model} \\
                Max sequence length &128&128&128&256&256&256 \\
                Vocabulary size &10,734 &61,040 &9513 &31,060 &8,259& 31,060  \\
                Batch size &32&128&32&64&32&64 \\
                Epochs &10&10&10&10&10&10& \\
                Learning rate &2e-5&3e-5&2e-5&2e-5&2e-5&2e-5 \\
                \hline
                {\bf DisBert/DisRoBERTa Model} \\
                Max sequence length &128&128&128&256&256&256 \\
                Batch size &32&128&32&32&32&64 \\
                Epochs &10&5&10&10&10&10& \\
                Learning rate &2e-5&3e-5&2e-5&3e-5&2e-5&3e-5 \\
                \hline
            \end{tabular}
            }
            \vspace{-3mm}
            \caption{Parameter settings.}  
        \label{parameter settings}
        \vspace{-3mm}
        \end{table*}   
\subsection{Heat Map}
\begin{figure}[htb]
        \centering
        \includegraphics[width=0.4\textwidth]{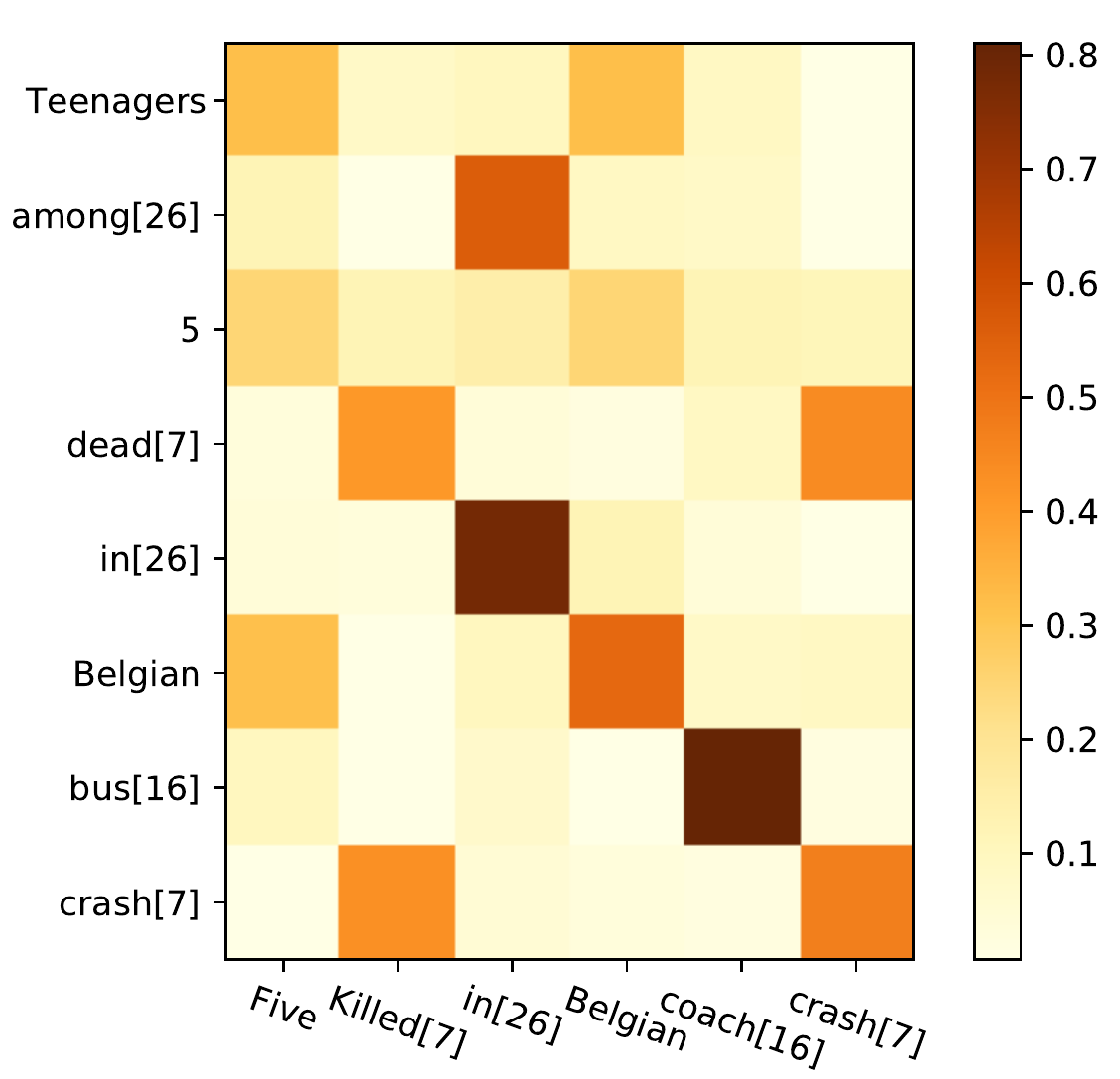}
        \caption{Heat Map calculated by semantics-driven multi-head attention.}
        \label{fig-heatmap}
    \end{figure}

    Figure \ref{fig-heatmap} shows the heat map of sentence pair 1 in the case study, which is calculated by our semantics-driven multi-head attention.
    We can observe that words from the same cluster usually pay high attention  to each other because they correspond to the same hidden vector in the codebook.
    Such attention could help Bert model focus on the words which may have similar semantics.
    
\label{heatmap}
    
\end{document}